\documentclass[]{interact}

\usepackage{epstopdf}% To incorporate .eps illustrations using PDFLaTeX, etc.
\usepackage{capt-of}
\usepackage[caption=false]{subfig}% Support for small, `sub' figures and tables
\usepackage[natbibapa,nodoi]{apacite}% Citation support using apacite.sty. Commands using natbib.sty MUST be deactivated first!
\theoremstyle{plain}% Theorem-like structures provided by amsthm.sty
\newtheorem{theorem}{Theorem}
\newtheorem{assumption}{Assumption}

\theoremstyle{definition}

\theoremstyle{remark}

\usepackage{mystyle}
\crefname{assumption}{Assumption}{Assumptions}

\begin{document}

% \articletype{ARTICLE TEMPLATE}% Specify the article type or omit as appropriate

\title{Fighting Selection Bias in Statistical Learning:\\
Application to Visual Recognition from Biased Image Databases}

\author{
\name{Stephan Cl\'{e}men\c{c}on\,\textsuperscript{a}, Pierre Laforgue\,\textsuperscript{b}\thanks{CONTACT Email: \href{mailto:pierre.laforgue1@gmail.com}{pierre.laforgue1@gmail.com}} and Robin Vogel\,\textsuperscript{c}}
\affil{
\textsuperscript{a} LTCI, T\'el\'ecom Paris, Institut Polytechnique de Paris, France\\
\textsuperscript{b} Universit\`{a} degli Studi di Milano, Milan, Italy\\
\textsuperscript{c} Monk AI, Paris, France}
}

\maketitle

\begin{abstract}
% Shorter version for ISNPS
In practice, and especially when training deep neural networks, visual recognition rules are often learned based on various sources of information. On the other hand, the recent deployment of facial recognition systems with uneven performances on different population segments has highlighted the representativeness issues induced by a naive aggregation of the datasets. In this paper, we show how biasing models can remedy these problems. Based on the (approximate) knowledge of the biasing mechanisms at work, our approach consists in reweighting the observations, so as to form a nearly debiased estimator of the target distribution. One key condition is that the supports of the biased distributions must partly overlap, and cover the support of the target distribution. In order to meet this requirement in practice, we propose to use a low dimensional image representation, shared across the image databases. Finally, we provide numerical experiments highlighting the relevance of our approach.

\end{abstract}

\begin{keywords}
Sampling bias; selection effect; visual recognition; reliable statistical learning
\end{keywords}

\section{Introduction}
Besides the considerable advances in memory technology and computational power, which now permit to implement optimization programs over vast classes of decision rules and make deep learning feasible, the spectacular rise in performance of visual recognition algorithms is essentially due to the recent availability of massive labeled image datasets.
However, due to the poor control of the data acquisition process, or even the absence of any experimental design to collect the datasets, the distribution of the training examples may drastically differs from that of the data to which the predictive rule will be applied when deployed.
One striking example of the adverse effects of selection bias in visual recognition is undoubtedly the racial bias recently identified and highlighted by various works on facial recognition systems.
Indeed, the largest publicly available face databases --- such as \cite{LFWTech} or \cite{y2016msceleb1m} for instance --- are composed of images of celebrities, and do not represent appropriately the global population, with respect to ethnicity and gender in particular.
In \cite{OtherFaceRecognition2}, face recognition systems learned from such databases are argued to suffer from an ``other-race effect'', i.e., that they fail to distinguish individuals from unfamiliar ethnicities, like humans often do.
Recently, industrial benchmarks for commercial face recognition algorithms highlighted the discrepancies in accuracy across race and gender, and their fairness is now questioned, see \cite{Grother2019}.
In \cite{racialfaces}, a balanced dataset over ethnicities
built by discarding selected observations from a larger dataset has been proposed so as to cope with the representativeness issue.
However, reducing the number of training samples is often an undesirable solution, as it may strongly damage performance and generalization for most computer vision tasks, see e.g., \cite{VinyalsBLKW16}.

In this paper, we show that provably accurate visual recognition algorithms can be learned from several biased image datasets, each of them possibly biased in a different manner with respect to the target distribution, as long as the biasing mechanisms are approximately known (or learned).
Whereas learning from training data that are not distributed ---or not of the same format--- as the test data is a very active research topic in machine learning, and in computer vision especially \citep{ASurveyTransferLearning,Ben-DavidBCKPV10}, with a plethora of transfer learning methods \citep{Tzeng2015} or domain adaptation techniques \citep{GaninJMLR}, the approach considered in this article is of different nature and relies on general user-defined biasing functions that relate the training datasets to the testing distribution.
In the facial recognition example detailed earlier, some typical biasing functions could leverage the gender and nationality information usually present in identification databases for instance.
However, we highlight that the approach promoted in this paper is very general, and applies whenever the specific types of selection bias present in the training sets are approximately known.
It is based on recent theoretical results on re-weighting techniques for learning from biased training samples, documented in \cite{laforgue2019statistical} (see also \cite{gill1988}) and \cite{vogel2020weighted}.

\begin{figure}[t]
\centering
\includegraphics[width=0.55\columnwidth]{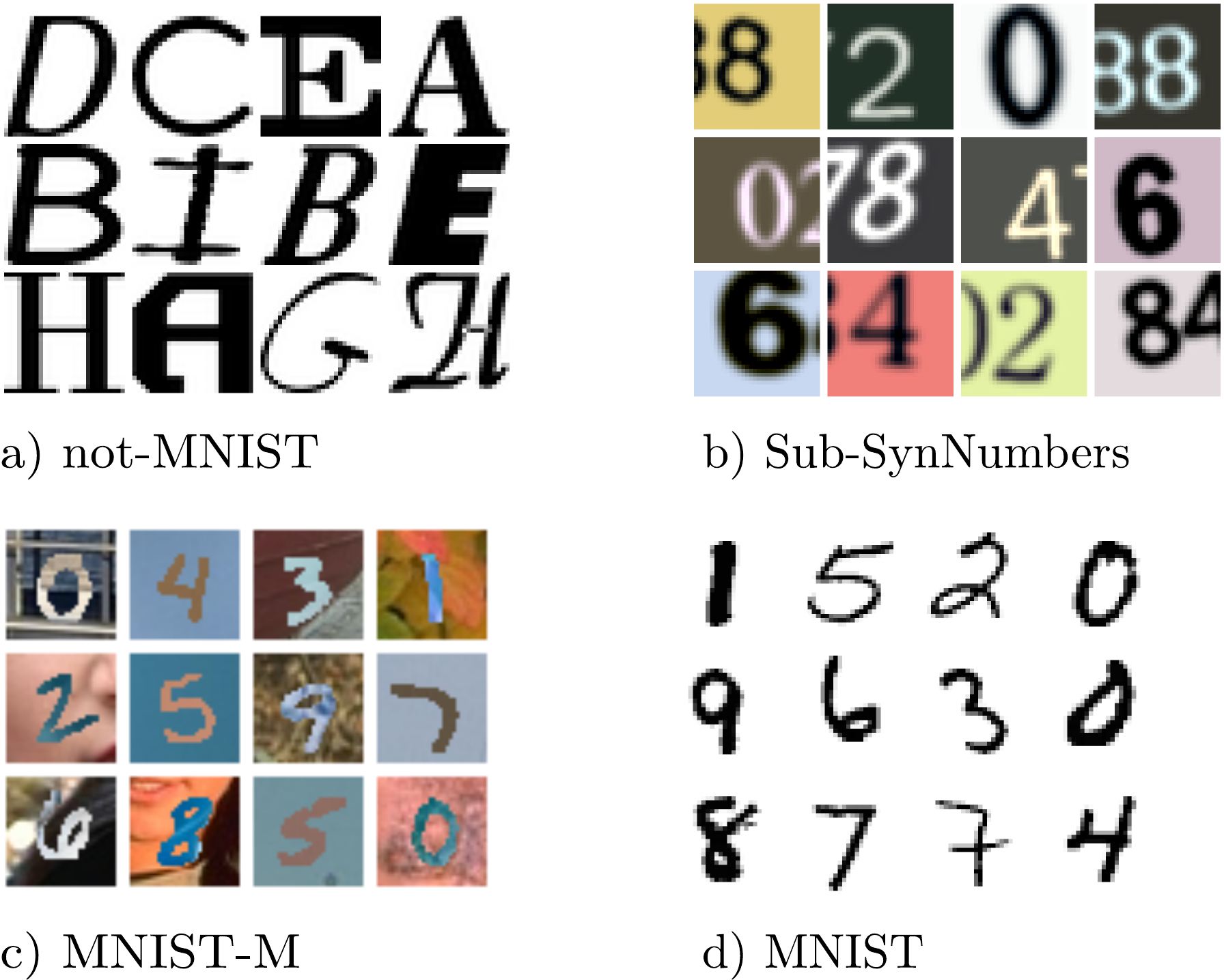}
\caption{A sample of: a) the not-MNIST dataset, b) the SynNumbers dataset of \cite{GaninJMLR} with only even numbers, c) the MNIST-M dataset of \cite{GaninJMLR}, and d) the well-known MNIST database. The difference between pairs of these datasets constitute examples of ``low-level'' bias (b - d, or c - d) and ``high-level'' bias (a - d, or b - d).}
\label{fig:three_types_bias}
\end{figure}

Note that the notion of \emph{bias} may carry different meanings.
In the facial recognition example, it is understood as a racial prejudice, due to the possibly great disparity between the performance levels attained for different subgroups of the population.
In computer vision, a significant part of the literature rather relates it to the case where the differences between the training and test image populations are due to low-level visual features \citep{LongZ0J16,BousmalisTSKE16,LiYSH18,GaninJMLR}, as illustrated for example by the difference between the MNIST and MNIST-M datasets, see \Cref{fig:three_types_bias}.
The general principle behind the algorithms proposed then consists in finding a representation common to training and test image datasets, that is relevant for the task considered.
In this paper, we focus on selection (or sampling) bias.
Namely, we consider situations where the training and test images have the same format but the distributions of the training image datasets available are different from the target distribution, i.e., the distribution of the statistical population of images to which the learned visual recognition system will be applied.
In the facial recognition example, selection bias results in different proportions of the race and gender categories among the train and test datasets.
In \Cref{fig:three_types_bias}, the difference of the underlying distribution between Sub-Synth (b) and MNIST (d) implies both different visual appearances and different distributions over the categories, since (b) contains a significantly higher proportion of even numbers than (d).

Our work is based on the debiasing procedure proposed in \cite{gill1988} and \cite{laforgue2019statistical}, that relies on the exact knowledge of the biasing functions.
By means of an appropriate algorithm, these functions permit to weight the observations of the training databases so as to form a nearly de-biased estimate of the test distribution.
We revisit this method from a practical perspective in the context of visual recognition, and show in particular that it is enough to approximately know (or learn) the biasing functions.
The latter are either fixed in advance based on prior information, or else estimated from auxiliary information related to the image data.
We further explain how this debiasing machinery can be practically used for visual recognition tasks, where the small overlap of the support of the biasing functions raises additional challenges.
We finally illustrate the relevance of our approach through various experiments on the well known datasets CIFAR10 and CIFAR100.
The rest of the paper is organized as follows.
In \Cref{sec:method}, we describe at length our methodology to debias image datasets. 
Experimental results providing empirical evidences of the advantages of our approach are then displayed in \Cref{sec:experiments}, while important related works are discussed in \Cref{sec:related_work}.
Some concluding remarks are gathered in \Cref{sec:conclusion}, and technical details are postponed to the Appendix section.

\section{Debiasing the Sources - Methodology}\label{sec:method}
\begin{figure*}[!ht]
    \centering
    \includegraphics[width=\textwidth]{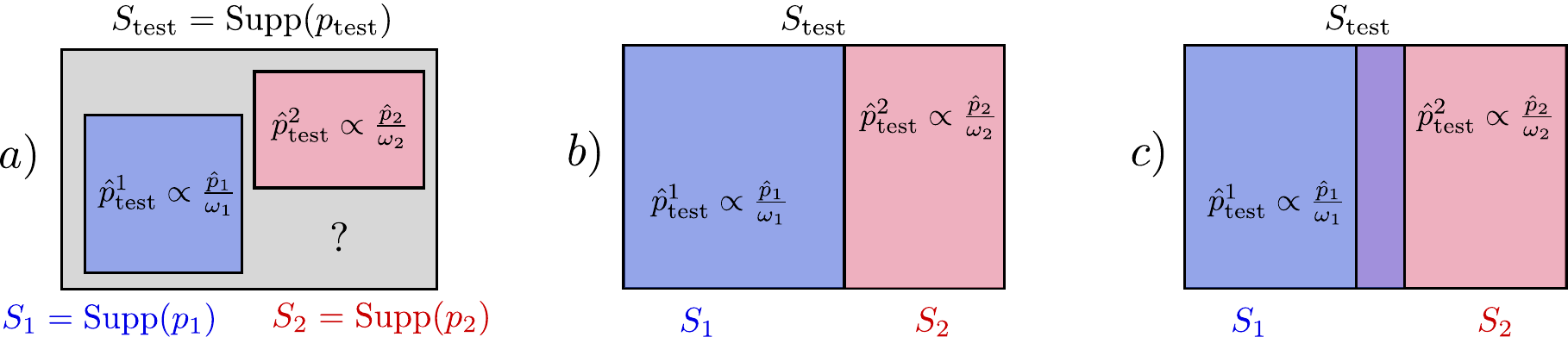}
    \caption{
    Given the datasets $\mathcal{D}_k$, it is easy to compute (unbiased) estimates of the $p_k$, through the empirical distributions $\hat{p}_k(z) = (1/n_k) \sum_{i=1}^{n_k} \mathbb{I}\{z = z_i^{(k)}\}$.
    Since the $\omega_k$ are assumed to be known, one can construct $\hat{p}_k/\omega_k$, normalize the distribution (to mimic the effect of $\Omega_k$ which is unknown), and obtain an estimate of $\ptest$ restricted to the support $S_k$ of $p_k$.
    But if the union of $S_1$ and $S_2$ is strictly included in $S_\text{test}$, as in case a), it is impossible to estimate $\ptest$ on $S_\text{test} \backslash(S_1 \cup S_2)$.
    This is expected, as the learner does not have access to any observation valued in this part of the space.
    If $S_\text{test}$ is included in $S_1 \cup S_2$, but there is no overlap, as in case b), we have estimates of $\ptest$ on the entire support, but it is impossible to find the right normalization to ensure that the combination of $\hat{p}_\text{test}^1$ and $\hat{p}_\text{test}^2$ is a good estimate of $\ptest$.
    More precisely, $0.1\hat{p}_\text{test}^1 + 0.9\hat{p}_\text{test}^2$ and $0.9\hat{p}_\text{test}^1 + 0.1\hat{p}_\text{test}^2$ are two distributions defined on $S_\text{test}$ that are equal to $\hat{p}_\text{test}^k$ if restricted to $S_k$, but from the data available in scenario b) there is no way to know which one is closer to $\ptest$.
    However, if there is overlap, as in case c), the knowledge of the $\omega_k$ and the observations present in the intersection allow to estimate the relative normalization, and to produce an almost unbiased estimate of $\ptest$.
    }
    \label{fig:explain-assumption}
\end{figure*}

In this section, we introduce our methodology to learn reliable decision functions in the context of biased image-generating sources.
In \Cref{subsec:db_erm}, we first recall the debiased Empirical Risk Minimization framework developed in \cite{laforgue2019statistical}, which our approach builds upon.
In \Cref{subsec:images}, we then highlight that applying this theory to image datasets raises important practical issues, and propose heuristics to bypass these difficulties.
We also show that an approximate knowledge of the biasing functions is sufficient, allowing for the design of an end-to-end method that jointly learns the biasing functions and the debiased model.
In what follows, we use $\mathbb{I}\{\cdot\}$ to denote the indicator of an event.

\subsection{Debiased Empirical Risk Minimization}
\label{subsec:db_erm}

Let $z = (x, y)$ be a generic random variable with distribution $p_\text{test}$ over the space $\mathcal{Z} = \mathcal{X} \times \mathcal{Y}$.
We consider the supervised learning framework, meaning that $x$ represents some input information (e.g., an image) supposedly useful to predict the output $y$ (e.g., the label of the image).
Given a loss function $\ell\colon \mathcal{Y} \times \mathcal{Y} \rightarrow \mathbb{R}_+$, and a hypothesis set $\mathcal{H} \subset \mathcal{Y}^\mathcal{X}$ (e.g., the set of decision functions possibly generated by a fixed neural architecture), our goal is to compute
\begin{equation}\label{eq:loss_expectation}
h^* = \argmin_{h \in \mathcal{H}} \, \mathcal{R}(h)\,, \qquad \text{where} \qquad \mathcal{R}(h) = \mathbb{E}_{(x, y) \sim \ptest}\big[\ell(h(x), y)\big]\,.
\end{equation}
As $\ptest$ is unknown in general, a usual hypothesis in statistical learning consists in assuming that the learner has access to a dataset composed of $n \in \mathbb{N}$ independent realizations generated from $\ptest$.
However, as highlighted in the introduction section, this i.i.d. assumption is often unrealistic in practice.
In this paper, we instead assume that the observations at disposal (the training images and labels) are generated by several biased sources.
Formally, let $K \in \mathbb{N}$ be the number of biased sources, and for $k \in \{1, \ldots, K\}$, let $\mathcal{D}_k = \big\{ z^{(k)}_1, \dots, z^{(k)}_{n_k}\big \}$ be a dataset of size $n_k$ composed of independent realizations generated from the biased distribution $p_k$.
We also define $n = \sum_{k=1}^K n_k$, $\lambda_k = n_k/n$, and $\underline{\lambda} = \min_k \lambda_k$.
The distributions $p_k$ are said to be \textit{biased}, as it is assumed the existence of biasing functions $\omega_k\colon \mathcal{Z} \rightarrow \mathbb{R}_+$ such that we have
\begin{align}\label{eq:bias-dist}
\forall\,z\in \mathcal{Z}, \qquad p_k(z) = \frac{\omega_k(z)}{ \Omega_k } \ptest(z)\,,
\end{align}
where $\Omega_k = \int  \omega_k(z) d\ptest(z)$ is a normalizing factor.
This data generation design is referred to as Biased Sampling Models (BSMs), and was originally introduced in \cite{vardi1985} and \cite{gill1988} for asymptotic nonparametric estimation of cumulative distribution functions.
Note that BSMs are a strict generalization of the standard statistical learning framework, insofar as the latter can be recovered as a special case with the choice $K=1$ and $\omega_1(z) = \mathbb{I}\{z \in \mathcal{Z}\}$.
For general BSMs, it is important to observe that performing naively Empirical Risk Minimization (ERM, see e.g., \cite{Devroye1996}), which means concatenating the $n$ observations without considering the sources of origin and computing
\begin{equation}\label{eq:loss_empirical}
\hat{h} = \argmin_{h \in \mathcal{H}} ~ \frac{1}{n}\sum_{k=1}^K\sum_{i=1}^{n_k} ~ \ell\Big(h\big(x_i^{(k)}\big), y_i^{(k)}\Big),
\end{equation}
might completely fail.
Indeed, solving \eqref{eq:loss_empirical} instead of \eqref{eq:loss_expectation} amounts to replace $\ptest$ with the empirical distribution
\[\hat{p}(z) = \frac{1}{n} \sum_{k=1}^K\sum_{i=1}^{n_k} \mathbb{I}\big\{z = z_i^{(k)}\big\} = \sum_{k=1}^K \lambda_k \hat{p}_k(z)\,, \quad~ \text{where} \quad~ \hat{p}_k(z) = \frac{1}{n_k} \sum_{i=1}^{n_k} \mathbb{I}\big\{z = z_i^{(k)}\big\}\,.
\]
However, it can be seen from the above equations that $\hat{p}$ is rather an approximation of the distribution $\bar{p} \coloneqq \sum_k \lambda_k p_k$, which might be very different from $p_\text{test}$.
For instance, in the extreme case where $\lambda_1 = 1$ and $\lambda_k = 0$ for $k \ne 1$, we have $\bar{p} = p_1 \ne p_\text{test}$.
To remedy this problem, \cite{laforgue2019statistical} has developed a debiased ERM procedure for BSMs under the following three (informally stated) assumptions: ($1$) the biasing functions $\omega_k$ are exactly known, ($2$) the union of the supports of the $p_k$ must contain the support of $\ptest$, and ($3$) the supports must \textit{sufficiently} overlap.
See \Cref{fig:explain-assumption} for an intuitive explanation of the assumptions.
Formally, the support assumption ensures that the $\omega_k$ cannot be null all at the same time, and thus allows to invert the following likelihood ratio.
Indeed, we have
\begin{equation}
\bar{p} = \sum_{k=1}^K \lambda_k p_k  = \sum_{k=1}^K \frac{\lambda_k\omega_k}{\Omega_k} \ptest\,, \qquad \text{such that} \qquad \ptest = \left(\sum_{k=1}^K \frac{\lambda_k\omega_k}{\Omega_k}\right)^{-1}\bar{p}\,.\label{eq:debias}
\end{equation}
As noticed earlier, it is straightforward to obtain an estimate of $\bar{p}$, by $\hat{p} = \sum_k \lambda_k \hat{p}_k$.
The $\omega_k$ being assumed to be known, it is then enough to find estimates of the $\Omega_k$, and to plug them into \eqref{eq:debias}, to construct an (almost unbiased) estimate of $\ptest$.
As shown in \cite{gill1988}, such estimates (denoted $\hat{W}_k$ in the following) can be easily computed by solving a system of $K-1$ equations through a Gradient Descent approach, see \Cref{apx:W} for technical details.
Let $\tilde{p}$ be the distribution obtained by replacing $\bar{p}$ and $\Omega_k$ in \eqref{eq:debias} with $\hat{p}$ and $\hat{W}_k$ respectively.
Debiased ERM consists in approximating $\ptest$ in \eqref{eq:loss_expectation} with $\tilde{p}$.
It can be shown that it is equivalent to compute
\begin{equation}\label{eq:decision_function}
\tilde{h} = \argmin_{h \in \mathcal{H}} ~ \sum_{k=1}^K\sum_{i=1}^{n_k} \pi_{ki} ~ \ell\Big(h\big(x_i^{(k)}\big), y_i^{(k)}\Big),
\end{equation}
where
\begin{equation}\label{eq:pi}
\pi_{ki} = \frac{\left(\sum_{l=1}^K \frac{\lambda_l \omega_l\big(z^{(k)}_i\big)}{\hat{W}_l}\right)^{-1}}{\sum_{m=1}^K\sum_{j=1}^{n_m}\left(\sum_{l=1}^K \frac{\lambda_l \omega_l\big(z^{(m)}_j\big)}{\hat{W}_l}\right)^{-1}}.
\end{equation}
Hence, once the debiasing weights $\pi_{ki}$ are computed, this approach boils down to weighted ERM.
Note also that most modern machine learning libraries allows to weight the training observations during the learning stage.%, e.g., through the \texttt{sample\_weight} keyword argument in the \texttt{fit} method for scikit-learn \citep{scikit-learn}.

\subsection{Application to Image-Generating Sources}
\label{subsec:images}

In this section, we detail how to apply the above framework to biased image-generating sources.
In particular, we make explicit the overlapping requirements, and propose a way to meet them when the input training data are images, which are known to be high-dimensional objects with non-overlapping supports.
We also relax the assumption that the biasing functions $\omega_k$ are exactly known, and show that similar guarantees can be achieved when conducting the procedure with some $(1/\sqrt{n})$-approximations of the $\omega_k$.
This result allows for the design of end-to-end debiasing procedure, where the biasing functions and the debiased decision rule are jointly learned.
\medskip

\par{\bf A framework well-suited to image datasets.}
By its generality, debiased ERM allows to model realistic scenarios, making its application to image-generating sources particularly relevant.
First recall that large image databases are often constructed by concatenating several datasets, each collected in different acquisition conditions.
This paradigm is precisely the one described by biased generating models, see \eqref{eq:bias-dist}.
Furthermore, note that the biasing functions $\omega_k$ apply to the entire observation $z = (x, y)$, covering for instance the typical scenario where the bias is due to different proportions of classes $y$ in the different datasets.
The latter mechanically occurs as soon as the datasets are collected by cameras located in different areas of the world, e.g., photographic safari in different countries will capture different species of animals, security cameras in the city or the countryside will record different types of vehicles.
This scenario is empirically addressed in \Cref{subsec:exp:bias_fun}.
But the model may also account for biases induced by the measurement devices used to collect the different datasets.
Indeed, cameras with different intrinsic
characteristics are expected to produce photos $x$ with different film grains.
This scenario is explored in \Cref{subsec:exp:est_bias}.
Of course, any combination of the above two kinds of bias is also covered.
This rich framework thus generalizes Covariate Shift \citep[e.g.]{quionero2009dataset,sugiyama2012machine}, a popular bias assumption which only allows the marginal distribution of $x$ to vary across the datasets, the conditional probability $p(y \mid x)$ being assumed to remain constant equal to $\ptest(y \mid x)$.

Another advantage of debiased ERM is that an individual biasing function $\omega_k$ might be null on some part of $\mathcal{Z}$.
This allows for instance to consider biasing functions of the form $\omega_k(z) = \mathbb{I}\{z \in \mathcal{Z}_k\}$, where $\mathcal{Z}_k$ is a subspace of $\mathcal{Z}$, accounting for the fact that dataset $\mathcal{D}_k$ is actually sampled from a subpopulation (e.g., regional animal species, as one cannot observe wild elephants in North America, light vehicles, since trucks are usually banned from downtown streets).
In this case, Importance Sampling typically fails, as the importance weights are given by $\ptest(z)/p_k(z) \propto 1 / \omega_k(z)$, which explodes for $z \notin \mathcal{Z}_k$.
In contrast, debiased ERM combines the different datasets in a clever fashion to produce an estimator valid on the entire set $\mathcal{Z}$.

Finally, we highlight that debiased ERM, unlike many \textit{ad hoc} debiasing heuristics, enjoy strong theoretical guarantees.
Indeed, it is shown in \cite{laforgue2019statistical} that, up to constant factors, $\tilde{h}$ has the same excess risk as an empirical risk minimizer that would have been computed on the basis of an unbiased dataset of size~$n$.
This remarkable property is however conditioned upon two relatively strong requirements, that we now study and relax in the context of image-generating sources.
\medskip

\begin{figure}[!t]
    \centering
    \includegraphics[width=0.55\columnwidth]{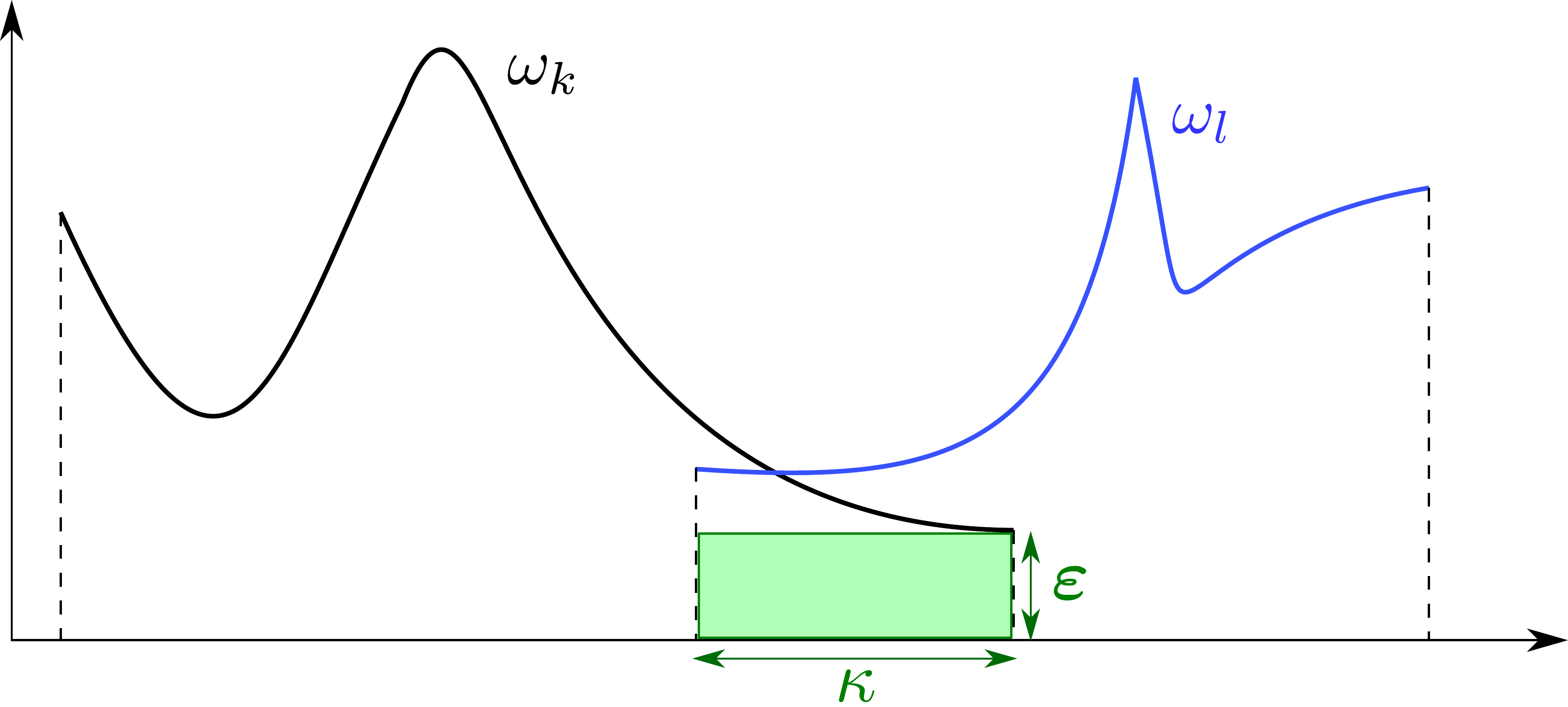}
    \caption{Two overlapping biasing functions.}
    \label{fig:overlap}
\end{figure}

\par{\bf Meeting the overlapping requirement.}
In this paragraph, we make explicit the overlapping requirements that are necessary to the application of debiased ERM, and propose a way to meet them in practice when dealing with image databases.
The two assumptions, reproduced from \cite{laforgue2019statistical}, are as follows.

\begin{assumption}\label{hyp:connect}
For $\kappa > 0$, define $G_\kappa$ the (undirected) graph with vertices in $\{1, \ldots, K\}$, and edge between $k$ and $l$ $(k\ne l)$ if and only if
\[
\int \mathbb{I}\{\omega_k(z) > 0 \} \cdot \mathbb{I}\{\omega_l(z) > 0\}d\ptest(z) \ge \kappa\,.
\]
There exists $\kappa > 0$ such that $G_\kappa$ is connected.
\end{assumption}

\begin{assumption}\label{hyp:bounded_omega}
There exists $\varepsilon > 0$ such that
\[
\forall\,z\in \mathcal{Z},~\forall\,k \in \{1, \ldots, K\}, \qquad \varepsilon\cdot\mathbb{I}\{\omega_k(z) > 0\} \le \omega_k(z) \le 1\,.
\]
\end{assumption}

In particular, \Cref{hyp:bounded_omega,hyp:connect} ensure that two overlapping biasing functions share at least a surface of $\kappa \cdot \varepsilon$, see \Cref{fig:overlap}.
An important caveat of the debiased ERM analysis is that constant factors heavily depend on this overlapping.
Indeed, results typically hold with probability $1 - e^{-cn}$, where $c$ is a constant proportional to $\kappa^2 \varepsilon^2$.
As previously discussed, we recover that theoretical guarantees are conditioned upon overlapping: when $\kappa =0$ or $\varepsilon = 0$, the bound obtained is vacuous.
But the formula actually suggests an even more interesting behavior.
Besides the $0$ threshold, we can see that the more distributions overlap, i.e., the bigger $\kappa$ and $\varepsilon$ are, the better the guarantees.
When the input observations are images, i.e., high dimensional objects whose distributions have non overlapping supports, it is likely that constants $\kappa$ and $\varepsilon$ become too small to provide a meaningful bound.
To remedy this practical issue, we propose to use biasing functions which take as input a low-dimensional embedding of the images (and/or the label $y$) in order to maximize the overlapping and thus ensure good constant factors.
We also conduct several sensitivity analyses, which corroborate the fact that increasing the density overlapping brings stability to the procedure, and globally enhances the performances, see e.g., \Cref{fig:dist_as_gamma}.
\medskip

\par{\bf Approximating the biasing functions.}
Another main limitation of debiased ERM is that the biasing functions $\omega_k$ are assumed to be exactly known.
Recall however that this assumption is less compelling that it might look like at first sight.
Indeed, only the $\omega_k$, and not the normalizing factors $\Omega_k$, are assumed to be known.
Consider again the case where $\omega_k(z) = \mathbb{I}\{z \in \mathcal{Z}_k\}$, with $\mathcal{Z}_k$ a subspace of $\mathcal{Z}$.
The assumption requires to know $\omega_k$, or equivalently the subpopulation $\mathcal{Z}_k$, but not to know the relative size of this subpopulation among the global one with respect to $\ptest$, i.e., $\int_{\mathcal{Z}_k}d\ptest(z) = \int_\mathcal{Z} \mathbb{I}\{z \in \mathcal{Z}_k\}d\ptest(z) = \Omega_k$, which is much more involved to obtain in practice.
Still, except in very particular cases such as randomized control trials, it is extremely rare to have an exact knowledge of the $\omega_k$.
In \Cref{thm:approx}, we show that using $(1/\sqrt{n})$-approximations of the biasing functions, denoted $\hat{\omega}_k$ in what follows, is actually enough to enjoy the theoretical guarantees of debiased ERM.

\begin{assumption}\label{hyp:approx_bias}
There exists a universal constant $C_\omega > 0$ such that for all $z \in \mathcal{Z}$ and any $k \in \{1, \ldots, K\}$, we have
\[0 \le \hat{\omega}_k(z) \le 1\,, \qquad \text{and} \qquad |\hat{\omega}_k(z) - \omega_k(z)| \le \frac{C_\omega}{\sqrt{n}}\,.\]
\end{assumption}

We also recall two assumptions from \citet{laforgue2019statistical}, that are necessary to derive guarantees for debiased ERM.

\begin{assumption}\label{hyp:eigenvalue}
Let $U = \log(2K/\varepsilon)\sum_{t=1}^{K-1}2^t(\underline{\lambda}\kappa \varepsilon)^{-t}$, $\mathcal{U} = [0, U]^K \subset \mathbb{R}^K$, and $\sigma > 0$.
For all $\bm{u} \in \mathcal{U}$, $\sigma_2(\bar{D}''(\bm{u})) \ge \sigma$, where $\sigma_2(A)$ denotes the second smallest eigenvalue of a matrix $A$, and $\bar{D}''(\bm{u})$ is defined in \eqref{eq:hessian}.
\end{assumption}

\begin{assumption}\label{hyp:complex}
The class of functions $\ell_\mathcal{H} = \big\{(x, y) \mapsto \ell\big(h(x), y\big) \colon h \in \mathcal{H} \big\}$ satisfies $\big|\ell\big(h(x), y\big)\big| \le 1$ for all $x, y, h$, and is a uniform Donsker class (relative to $L_2$) with polynomial uniform covering numbers, i.e., there exist constants $C_\mathcal{H}>0$ and $r\geq 1$ such that for all $\zeta>0$
\[
\sup_{Q}\,\mathcal{N}(\zeta,\; \ell_\mathcal{H},\; L_2(Q) )\leq (C_\mathcal{H}/\zeta)^r\,,
\]
where the supremum is taken over the probability measures on $\mathcal{Z}$, and $\mathcal{N}(\zeta, \ell_\mathcal{H}, L_2(Q) )$ is the minimum number of $L_2(Q)$ balls of radius $\zeta$ needed to cover $\ell_\mathcal{H}$.
\end{assumption}

As can be seen in \Cref{apx:W}, debiased ERM builds on the resolution of a system of equations to compute some estimates of the $\Omega_k$.
\Cref{hyp:eigenvalue} is a technical condition which ensures that this system is well-behaved.
\Cref{hyp:complex} is a standard complexity assumption, see e.g., \citet{VanderVaart1996}.
Assuming that one has access to estimates $\hat{\omega}_k$ of the bias functions $\omega_k$, we can state the guarantees of running debiased ERM with the $\hat{\omega}_k$ instead of the $\omega_k$.

\begin{theorem}\label{thm:approx}
Suppose that \Cref{hyp:connect,hyp:bounded_omega,hyp:approx_bias,hyp:eigenvalue,hyp:complex} are satisfied.
Let $\tilde{h}_\mathrm{approx}$ be the decision function computed from the debiased ERM procedure, see \eqref{eq:decision_function}, where the $\hat{\omega}_k$ are used instead of the $\omega_k$.
Then, there exist constants $M, M', M'', c, c', c'', \gamma, n_0$, depending only on $K, \underline{\lambda}, \kappa, \varepsilon, C_\omega, \sigma, C_\mathcal{H}, r$, such that for all $t > 0$ and $n \ge n_0$ it holds
\[
\mathbb{P}\left( \mathcal{R}(\tilde{h}_\mathrm{approx}) - \mathcal{R}(h^*) > \frac{\gamma}{\sqrt{n}} + t\right) \le Me^{-cn} + M'e^{-c'nt^2} + (\sqrt{n}t)^r M''e^{-c''nt^2}\,.
\]
\end{theorem}

\Cref{thm:approx} is fundamental, as it shows that using approximations of the $\omega_k$, for instance learned from the data, is enough to achieve small excess of risk in biased sampling models.
Our approach thus provides an end-to-end methodology to debias image-generating sources, which only relies on the biased databases, does not use any oracle quantities (the $\omega_k$), and can therefore be fully implemented in practice.
Note that if one has an idea on the feature responsible for the bias (which is very common in practice), then some $\hat{\omega}_k$ can easily be computed by estimating the right thresholds, see \Cref{sec:experiments}.
These procedures are known to converge at a $1/\sqrt{n}$ rate, and therefore fulfill \Cref{hyp:approx_bias}.

\section{Experiments}\label{sec:experiments}
We now present experimental results on two standard
image recognition datasets, namely CIFAR10 and CIFAR100,
% where the training dataset is split to form $K$ biased datasets so as to illustrate \cref{sec:method}.
from which we extract $K$ biased datasets, so as to recreate the framework of \Cref{subsec:db_erm} in a controllable way.
% While \cite{laforgue2019statistical} proposes several settings where the bias functions $\omega_k$'s are specified over the whole input space (\textit{e.g.} with censored data), their application to image recognition scenarios is not straightforward. Indeed, as discussed in \cref{sec:method}, biasing functions cannot in general be expressed as a function over the image space due to its dimensionality.
% However, as discussed in \Cref{subsec:images}, applying debiased ERM to image recognition problems is not straightforward. For instance, modelling the bias through functions $\omega_k$ naively defined on the inputs is impossible: the raw images being high dimensional objects, the overlapping condition won't be satisfied, and debiased ERM impossible to compute. Thus, we have to resort to biasing models based on more compact representations the data, such as HSV, or which play directly on the class $y$.
The great generality of this setting allows us to model realistic scenarios, such as the following.
An image database is actually composed of pictures that have been collected in several goes, by cameras located in different parts of the world.
One may think about animal pictures gathered by expeditions in different countries, or security cameras located in different areas for instance.
In this case, the location bias then translates into a class bias, and the class proportions greatly differ from one database to the other (we do not observe the same species of animals in Africa or in Europe).
We show that our method allows to efficiently approximate the biasing functions in this case, and provides a satisfactory end-to-end solution to this common bias scenario.
%
% In this section, we considered two natural settings where the databases $\mathcal{D}_1, \dots, \mathcal{D}_K$ can be debiased although the bias functions are only approximately known, showing the relevance of the approach.
% Moreover, \cite{laforgue2019statistical} requires an exact knowledge of the biasing functions $\omega_k$, which is often unrealistic in practice. In this section, we consider two natural settings where the biasing functions are learned as a part of the learning process. The good empirical performances we obtain endorse the relevance of the approach.
A second typical bias scenario occurs when the images originate from cameras with different characteristics, and thus exhibit different film grains.
We show that a bias model on a low representation of the images (therefore avoiding the overlapping difficulty) is able to model such phenomenon, and that our approach provides satisfactory results despite the bias mechanism at work.

\textit{Imbalanced learning} is a single dataset image recognition problem, in which the training set is assumed to contain the classes of interest in a proportion different from the one in the validation set (usually supposed balanced).
Long-tail learning, a special instance of imbalanced learning, has recently received a great deal of attention in the computer vision community, see e.g., \cite{MenonJRJVK21}.
In \Cref{subsec:exp:bias_fun}, we propose a multi-dataset generalization of imbalanced learning, where we assume that each database $\mathcal{D}_k$ contains a different proportion of the classes of interest, this proportion being possibly equal to $0$ for certain classes (which is typically not allowed in single dataset imbalanced learning).
Under the assumption that the classes are balanced in the validation dataset, we show that $\hat{\omega}_k(z) = \hat{\omega}_k(x, y) = n_k^{(y)}$, where $n_k^{(y)}$ is the number of observations with label $y$ in dataset $\mathcal{D}_k$, approximates well the bias function.
The good empirical results we obtain provide a proof of concept for our methodology in image recognition problems where the marginal distributions over classes differ between the data acquisition phase and the testing stage.
Note that this scenario is often found in practice, e.g., as soon as cameras are located in different places of the world and thus record objects in proportions which are specific to their locations (more scooters in the city,
tractors in the countryside, ...).
Recall that our goal here is not to achieve the best possible accuracy, otherwise knowing the test proportions is enough to reweight the observations accurately, but rather to show that our method achieves reasonable performances, even without knowledge of the $\omega_k$.
We also conduct a sensitivity analysis, showing that increasing the overlapping indeed improves
stability and accuracy.

In \Cref{subsec:exp:est_bias}, we assume that the biased datasets $\mathcal{D}_k$ are collected under different acquisition conditions (e.g., quality of the camera used, scene illumination, ...), thus generating different types of images.
To approximate the bias function, we first embed the images
into a small space where the different types are easily separable.
Then, we use the available instances in $\mathcal{D}_k$ to estimate the boundaries between the different image types.
This experimental design complement the previous one, as bias now apply to (a transformation of) the inputs $x$, and not on the classes $y$, thus highlighting the versatility of our approach in terms of bias that can be modeled.
In our experiments, we consider a scenario where images have distinct backgrounds, but of course the debiasing technique could be applied to any other intrinsic property of the images, such as the illumination of the scene.

\subsection{Experimental Details}
\label{subsec:exp:expe_details}

CIFAR10 and CIFAR100 are two standard image recognition datasets, that contain both 50K training images and 10K testing images, of resolution $32\times32$ pixels.
CIFAR10 is a 10-way classification dataset, while CIFAR100 is a 100-way classification dataset.
Both the training sets and the testing sets are balanced in terms of the number of images per class.
CIFAR10 (resp. CIFAR100) thus contains 5K (resp. 500) images per class in the training split and 1K (resp. 100) images per class in the testing split.
Our experimental protocol consists of four steps:
\begin{enumerate}
\item we create the biased datasets $\mathcal{D}_k$ by
sampling observations with replacement (and according to the true $\omega_k$) from the train split of CIFAR10 (resp. 100)\,,
\smallskip
\item we construct estimates $\hat{\omega}_k$ of the biasing function $\omega_k$, based on the information contained in the datasets $\mathcal{D}_k$, and using any additional information on the testing distribution if available\,,\smallskip
\item we compute the $\hat{W}_k$ according to the procedure detailed in the Appendix, and derive the debiasing weights $\pi_{ki}$, see \eqref{eq:pi}, where we use $\hat{\omega}_k$ instead of $\omega_k$\,,\smallskip
\item we learn a model from scratch using the debiasing weights computed in step 3\,.
\end{enumerate}

Steps 1 and 2 are setting-dependent, as they use respectively an exact or incomplete knowledge about the bias mechanism at work.
The exact expression is used in step~1 to generate the training databases from the original training split.
It is however not used in the subsequent learning procedure.
The approximate expression serves as a prior in step 2 to produce the estimates $\hat{\omega}_k$.
The latter are typically determined by estimating from the databases $\mathcal{D}_1, \ldots, \mathcal{D}_K$ the value of an unknown parameter in the approximate expression.
Choices for the approximate expression can be motivated by expert knowledge, possibly combined with a preliminary analysis of the biased databases.
We provide a detailed description of steps 1 and 2 for each experiment in their respective sections.

In step 3, we compute the $\hat{W}_k$, which according to the computations in the Appendix amounts to minimize over $\mathbb{R}^K$ the (strongly) convex function $\widehat{D}$ defined in \eqref{eq:min_D}.
As $\widehat{D}$ is separable in the observations, we use the pytorch \citep{pytorch} implementation of Stochastic Gradient Descent (SGD) for $4000$ iterations, with batch size $100$, fixed learning rate $10^{-2}$, and momentum $0.9$.

In step 4, we train the ResNet56 implementation of \cite{He2015} for 205 epochs using SGD with fixed initial learning rate $10^{-1}$ and momentum $0.9$, where we divide the learning rate by $10$ at epoch $103$ and $154$.
Our implementation is available at \url{https://tinyurl.com/5ahadh7c}.

\subsection{Class Imbalance Scenario}
\label{subsec:exp:bias_fun}

In this section, we assume that the conditional probability of $x$ given $y$ is the same for all the distributions $p_1, \dots p_K, p_{\text{test}}$.
Formally, for all $y \in \mathcal{Y}$, and any $k \in \{1, \ldots, K\}$, we have $p_k(x \mid y) = \ptest(x \mid y)$.
However, the class probabilities $p_k(y)$ may vary from one distribution to the other, and we recover the link to imbalanced learning.
This is a common scenario, already studied in \cite{vogel2020weighted} for instance.
In this context, it can be shown that the biasing function on $z = (x, y)$ is proportional to the likelihood ratio between the class proportion of the biased dataset $p_k(y)$ and that of the testing distribution $p_{\text{test}}(y)$.
Indeed, using \eqref{eq:bias-dist} we obtain
\begin{align}\label{eq:bias_fun_imbalanced_classif}
\omega_k(x, y) = \Omega_k \frac{p_k(x,y)}{p_{\text{test}}(x,y)} = \Omega_k \frac{p_k(y)}{p_{\text{test}}(y)}\,.
\end{align}
Since we assume the classes to be balanced in the test set, we also have $p_{\text{test}}(y) \equiv 1/|\mathcal{Y}|$.
It then follows from \eqref{eq:bias_fun_imbalanced_classif}
that knowing $p_k(y)$ gives $\omega_k(z)$, up to a constant.
Similarly, substituting in \eqref{eq:bias_fun_imbalanced_classif} an estimate of $p_k(y)$ based on $\mathcal{D}_k$ immediately provides an estimate of $\omega_k$.
Although our experiment considers the classes associated to the datapoints as subjects of the bias, note that the knowledge of any discrete attribute of the data can be used for debiaising as long as both: 1) realizations of that attribute are known for the training set, and 2) its marginal distribution on the testing distribution is known.
Those additional discrete attributes are usually referred to as \textit{strata}.
\medskip

\par
\begin{figure}[t]
    \centering
    \includegraphics[width=0.8\columnwidth]{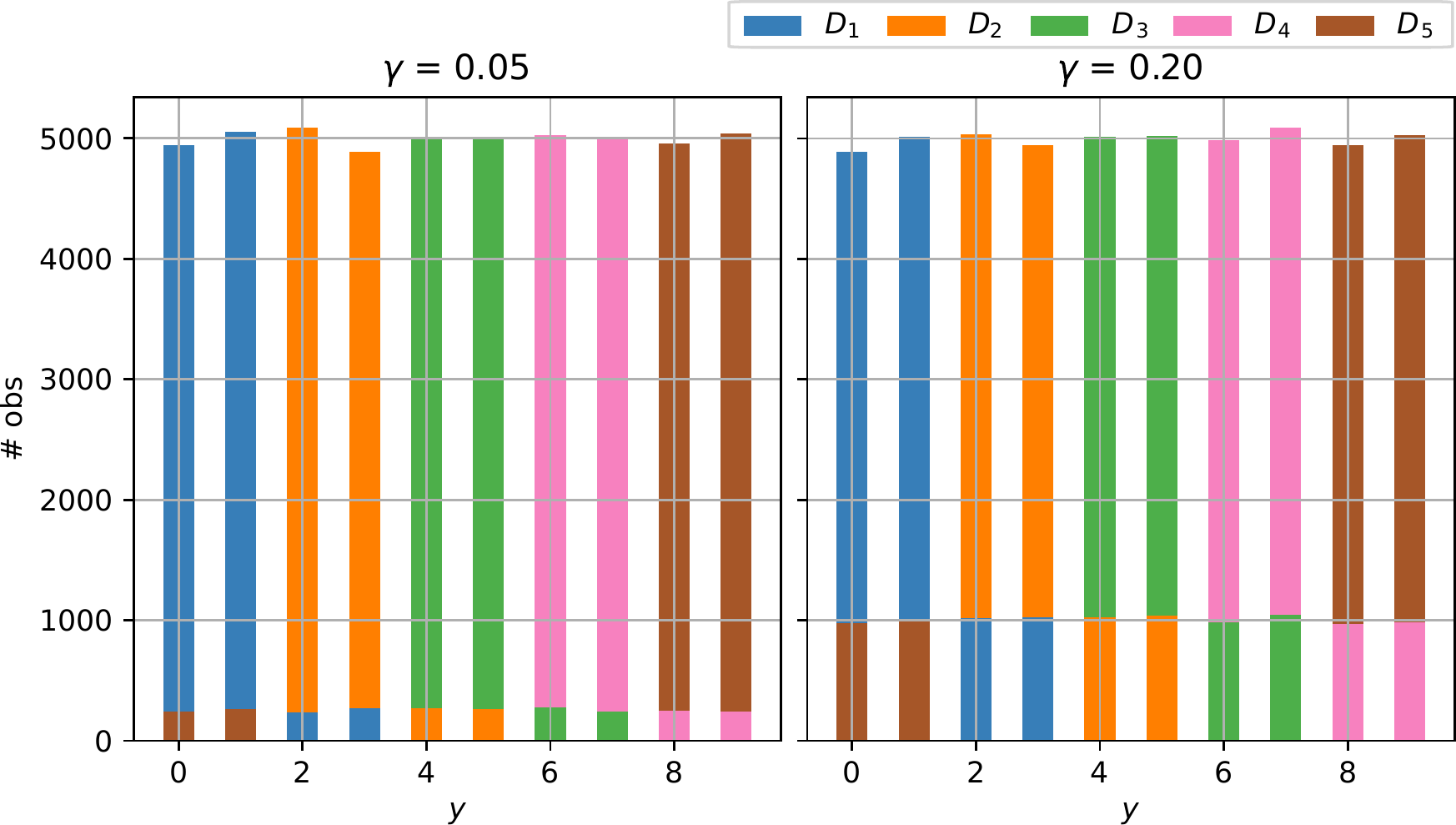}
    \caption{Number of element per class for the datasets $\mathcal{D}_1,\dots,\mathcal{D}_5$
	in a scenario where $K = 5$ and $\gamma$ either low (0.05) or high (0.2) for CIFAR10.
    }\label{fig:synthetic_example}
\end{figure}

{\bf Generation of the biased datasets.}
Let $M = |\mathcal{Y}|$, and choose $K$ which divides $M$ (here we take $K=5$).
We split the original train dataset into $K$ subsamples with different class proportions
as follows.
We first partition the classes into $K$ meta-classes $\mathcal{C}_1, \ldots, \mathcal{C}_K$, such that each $\mathcal{C}_k$ contains $n_\mathcal{C} \coloneqq M/K$ different classes.
In the case of CIFAR-10 we have $\mathcal{C}_1 = \{0, 1\}, \mathcal{C}_2 = \{2, 3\}, \ldots, \mathcal{C}_5 = \{8, 9\}$.
Then, we consider that each dataset $\mathcal{D}_k$ is composed of classes in meta-classes $\mathcal{C}_k$ and $\mathcal{C}_{k+1}$ only, and that classes coming from the same meta-class are in the same proportion.
Formally, for $k \in \{1, \ldots, K\}$, we have\vspace{-0.2cm}
\[
p_k(y) =
\begin{cases}
\frac{1 -\gamma}{n_\mathcal{C}} & \text{ if } y \in \mathcal{C}_k\\
\frac{\gamma}{n_\mathcal{C}} & \text{ if } y \in \mathcal{C}_{k+1}\\
0 & \text{ otherwise}
\end{cases}
\]
with the convention $\mathcal{C}_{K+1} = \mathcal{C}_1$,
and where $\gamma \in [0, 0.5]$ is an overlap parameter.
Note that the concatenation of the datasets $\mathcal{D}_1, \dots, \mathcal{D}_K$ has the same distribution over the classes (in expectation) as the original unbiased training dataset.
A classifier learned on this concatenation will therefore serve as a reference.
Note also that some observations of the original dataset may not be present in the resampled dataset, which is a consequence of sampling with replacement.
For $K=5$, we illustrate the distribution over classes of all datasets $\mathcal{D}_1, \dots, \mathcal{D}_5$ for different values of $\gamma$ in \Cref{fig:synthetic_example}.
\medskip

\par{\bf Approximating the biasing functions.}
Let $n_k^{(y)}$ be the number of observations with class $y$ in database $\mathcal{D}_k$.
A natural approximation of $p_k(y)$ is $n_k^{(y)}/n_k$.
Furthermore, with the above construction of the biased datasets, we know that $\Omega_k$ and $n_k$ are the same for all $k \in \{1, \ldots, K\}$, and that $\ptest(y) = 1/M$.
Substituting these values in \Cref{eq:bias_fun_imbalanced_classif}, and since it is enough to know the $\omega_k$ up to a constant multiplicative factor (see \Cref{subsec:db_erm}), a natural approximation is
\[
\hat{\omega}_k(y) = n_k^{(y)}\,.
\]
We then solve for the debiasing sampling weights $\pi_{k, i}$ by plugging the approximations $\hat{\omega}_k$'s into the procedure described in \Cref{sec:method}.
\medskip

\par
\begin{figure*}[t]
    \centering
    \includegraphics[width=0.95\linewidth]{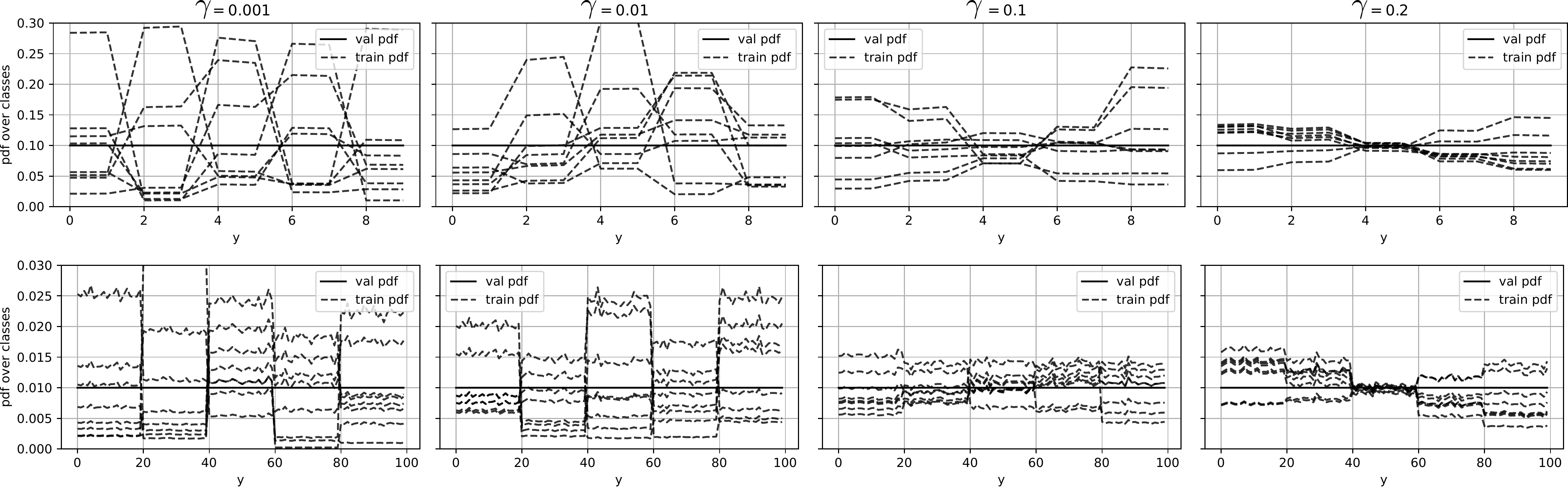}
    \caption{Distribution over the classes in the bias-corrected samples (dashed line, 8 different runs), compared to the testing distribution (full line) for both CIFAR10 (top) and CIFAR100 (bottom) and different overlap parameters $\gamma$ (left to right).}
    \label{fig:dist_against_gt}
    \vspace{.6cm}
\end{figure*}

\begin{figure}
\centering
\begin{minipage}[c]{0.45\textwidth}
\centering
\includegraphics[width=\columnwidth]{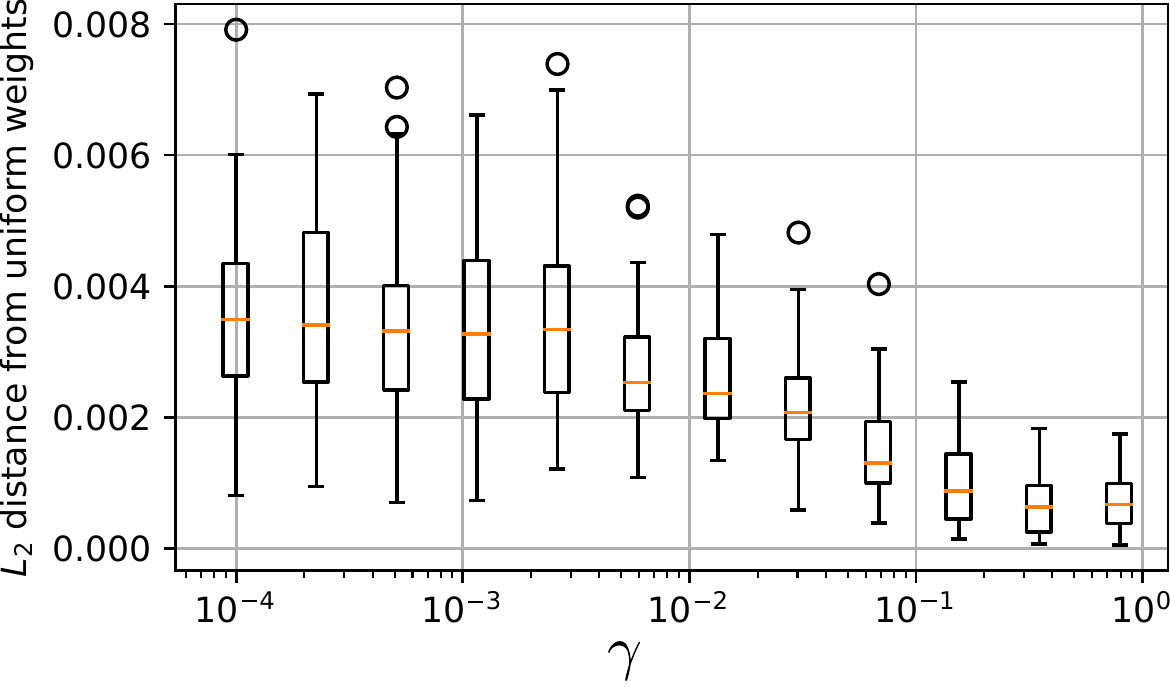}
\caption{Expected $L_2$ distance of the weights as a function of $\gamma$ for the CIFAR10 and $K=5$. The expected $L_2$ distance decreases as $\gamma$ grows.}
\label{fig:dist_as_gamma}
\end{minipage}
\hfill
\begin{minipage}[c]{0.49\textwidth}
\centering
\begin{scriptsize}
\begin{tabular}{cc}
\toprule
\multicolumn{2}{c}{CIFAR10} \\
Ref & $92.93 \; (\pm 0.47)$ \\
\midrule
$\gamma$ & Acc. \\
\midrule
$10^{-3}$ & $89.96 \;(\pm 2.15)$ \\
$10^{-2}$ & $90.51 \;(\pm 1.72)$ \\
0.10 & $90.97 \; (\pm 1.64)$ \\
0.20 & $91.02 \; (\pm 1.64)$ \\
\bottomrule
\end{tabular}
\quad
\begin{tabular}{cc}
\toprule
\multicolumn{2}{c}{CIFAR100} \\
Ref & $69.77 \;(\pm 1.12)$ \\
\midrule
$\gamma$ & Acc. \\
\midrule
$10^{-3}$ & $62.00 \; (\pm 6.73)$ \\
$10^{-2}$ & $63.66 \; (\pm 3.66)$ \\
0.10 & $65.08 \; (\pm 0.74)$ \\
0.20 & $64.96 \; (\pm 1.96)$ \\
\bottomrule
\end{tabular}
\end{scriptsize}
\medskip
\captionof{table}{Accuracy for different values of $\gamma$ on CIFAR for the experiments of \Cref{subsec:exp:bias_fun}, with 2 times the standard deviation over 8 runs between parenthesis. Learning with the original training dataset of CIFAR gives the Ref accuracy.
}
\label{tab:table_accs}
\end{minipage}
\end{figure}

{\bf Analysis of the debiasing weights.}
In \Cref{fig:dist_against_gt}, we plot the distribution 
over the classes in the datasets weighted by the debiaising procedure, and compare it to the test distribution over the classes, for 8 different runs of the debiasing procedure.
We observe that the distributions are significantly further
from the test distribution and more noisy as $\gamma$ decreases.
If we concatenated the databases $\mathcal{D}_1, \dots, \mathcal{D}_K$ into a single training dataset, i.e., set $\pi_{i,k}$ constant, then the concatenated dataset is equal to the original training dataset of CIFAR in expectation with respect to our synthetic biased generation procedure.
Therefore, it is desirable to obtain uniform weights after solving for the debiaising procedure.
In \Cref{fig:dist_as_gamma}, we plot the $\mathcal{L}_2$
distance between the obtained sampling weights and the uniform weights, as a function of $\gamma$.
The result shows that the distance from the optimal weights is higher when $\gamma$ is low and lower when $\gamma$ is high, confirming the overlap assumptions introduced
in \Cref{sec:method}.
\medskip

\par{\bf Empirical results.}
We provide testing accuracies for several $\gamma$ in \Cref{tab:table_accs}.
A slight performance decrease is observed in both cases, stronger in the case of CIFAR100.
We see that the predictive performance increases as $\gamma$ grows, while our estimate over 8 runs of the variance of the final accuracy decreases.
Our results confirm empirically the necessity to have a significant enough overlap between the databases.

\subsection{Image Acquisition Scenario}
\label{subsec:exp:est_bias}

In many cases, the bias does not originate from class
imbalances, but from an intrinsic property of the images,
such as the illumination of the scene or the background.
In this image acquisition scenario, we simulate the effect of learning with datasets collected on different backgrounds.
\medskip

\par
\begin{figure*}[t]
    \centering
    \includegraphics[width=0.95\linewidth]{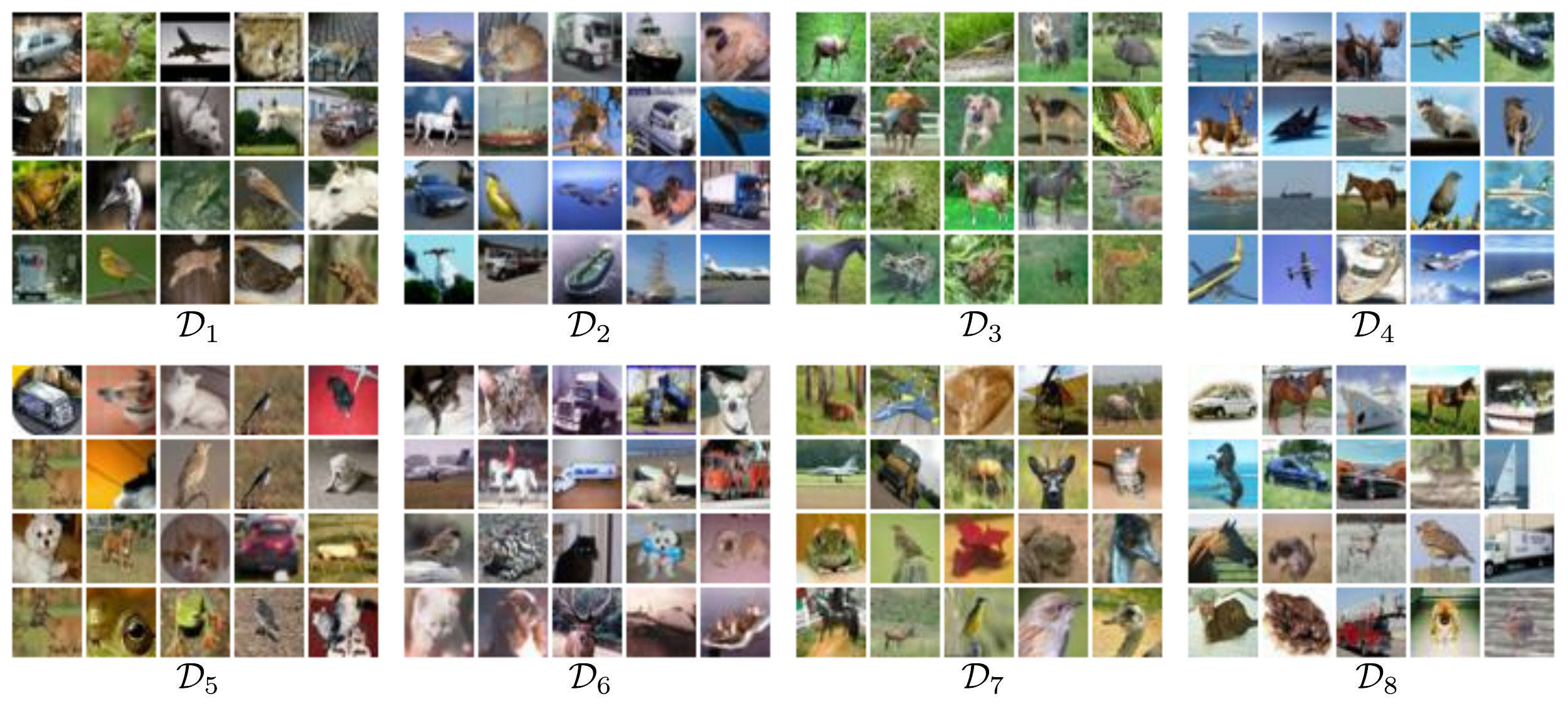}
    \caption{Examples of images obtained by separating the dataset CIFAR10 
	using the median thresholds on the average HSV values of the 2-pixel borders of the images.}
    \label{fig:samples_cifar10_hsv_sep}
\end{figure*}

{\bf Generation of the biased datasets.}
In CIFAR10 and CIFAR100, the background varies significantly between images.
Some images feature a textured background while others have an uniform saturated background.
We generate the biased datasets by grouping the images into bins derived from the average parametric color value in normalized HSV representation of a 2 pixel border around the image.
For an image $x$, we denote this 3-dimension representation of an image by $\widebar{x} = (\widebar{x}_H, \widebar{x}_S, \widebar{x}_V) \in [0, 1]^3$.
Each coordinate in this 3-dimensional space is then split using the median values $h_H, h_S, h_V$ of the encodings $\widebar{x}$ of the training dataset, giving $8$ bins $B_0, \dots, B_7 \subset [0, 1]^3$ to split the training images of CIFAR into 8 datasets $\mathcal{D}_1, \dots, \mathcal{D}_8$, see \Cref{fig:samples_cifar10_hsv_sep}.
Precisely, for any $l \in \{ 0, \dots, 7 \}$, if $(b_H, b_S, b_V) \in \{ 0, 1\}^3$ is the binary representation of $l$, i.e., $l = 4 b_H + 2b_S + b_V$, then
\[
B_l = \Big\{ \bar{x} \in [0, 1]^3 \colon \forall C \in \{ H, S, V \}, \left(\widebar{x}_C -h_C \right) \left(2 b_C - 1\right) \ge 0 \Big\}\,.
\]
To satisfy the overlap condition, we define the biasing function $\omega_k(x)$ for each image $x$ and any $k \in \{1, \dots, 8\}$, as
\[
\omega_k(x) = 
\begin{cases}
1 & \text{ if } \widebar{x} \in B_k,\\
\max \left( 0, 1 - \gamma^{-1} \norm{\widebar{x} - \Pi_1(\widebar{x}, B_k) }_1 \right) & \text{ if } \widebar{x} \notin B_k,
\end{cases}
\]
where $\Pi_1(\widebar{x}, B_k)$ is the projection of $\widebar{x}$ on $B_k$ with respect to the $\ell_1$ norm.
We sample a fixed number of elements per dataset $\mathcal{D}_k$ with possible redundancy (due to sampling with replacement) using the biased functions $\omega_k$ as sampling weights.
\medskip

\begin{figure*}[!t]
    \centering
    \includegraphics[width=\linewidth]{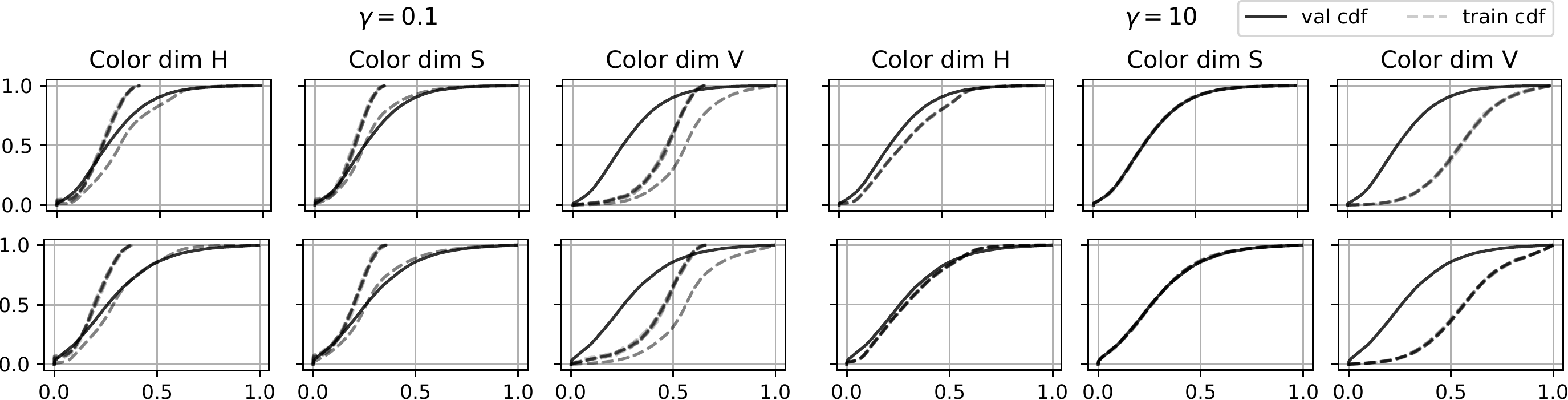}
    \caption{Distribution over the HSV dimensions of the average 2-pixel borders
	of the bias-corrected samples (dashed line, 8 different samples)
	compared to the testing distribution (full line),
	for both CIFAR10 (top) and CIFAR100 (bottom) and different overlap parameters
	$\gamma$ (left to right).}\label{fig:dist_over_bins}
\end{figure*}

\par{\bf Approximating the biasing functions.}
We first compute the minimum $m^{(k)} \in [0, 1]^3$ and maximum $M^{(k)} \in [0, 1]^3$ of each datasets $\mathcal{D}_k$ for the average border values $\widebar{x}$,
with
\[
m^{(k)} = \left(m^{(k)}_H, m^{(k)}_S, m^{(k)}_V\right)\,, \quad \text{and} \quad M^{(k)} = \left(M^{(k)}_H, M^{(k)}_S, M^{(k)}_V\right)\,.
\]
Then, we approximate the biasing functions by
\begin{align}\label{eq:approx_bias_expe2}
\hat{\omega}_k(x) = \I\left\{ \widebar{x}_C \in [m_C^{(k)}, M_C^{(k)}] \text{ for all } C \in \{H, S, V\} \right\}\,.
\end{align}
\medskip

\par{\bf Balanced datasets scenario.}
We first considered a balanced scenario, where each dataset
has the same number of instances, \textit{i.e.} $n_k = 50,000/K$ for all $k \in \{1, \dots, K\}$.
In this scenario, the debiased dataset is still biased in some sense, since \eqref{eq:approx_bias_expe2} does not take into account the natural distribution of the CIFAR training dataset onto the bins $B_0, \dots, B_7$ (see top figure of \Cref{fig:dist_over_bins}).
Although the results should improve with the debiasing procedure, they cannot be compared directly to the classification accuracy with the original training dataset of CIFAR.
Following the outline of \Cref{subsec:exp:bias_fun},
we provide in \Cref{fig:dist_over_bins} the marginal distribution of the bias-corrected samples over the HSV dimensions of the average 2-pixel borders and compare them to the test marginal distributions.
The corrected distributions are closer to the true
distributions and less random when $\gamma$  is high, although some expected bias remains.
We provide the corresponding accuracies as well as
an estimator of their variance for the learned models in \Cref{tab:table_accs_expe2}.
Our results show that the accuracy of the debiasing procedure increases with the overlap.
\medskip

\begin{figure}[t]
    \centering
    \includegraphics[width=\linewidth]{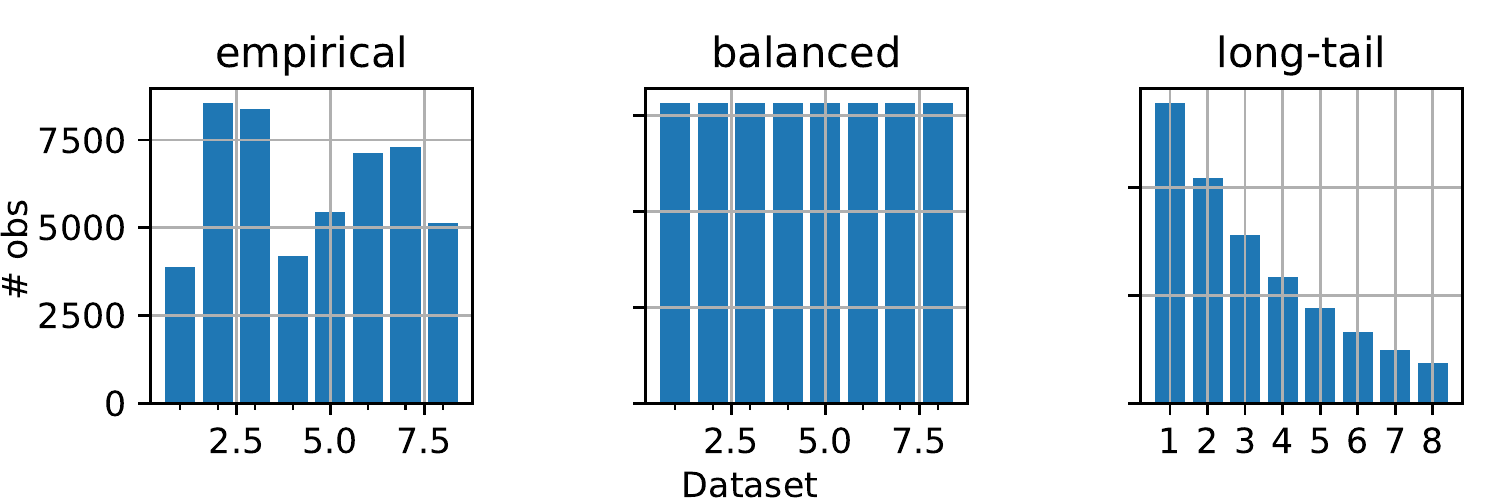}
    \caption{Distribution over the bins $B_0, \dots, B_7$
	of the training data (left), of the balanced (middle)
	and of the imbalanced scenario (right).
    }\label{fig:bins_cifar10}
\end{figure}

\begin{figure}
\centering
\begin{minipage}[c]{0.48\textwidth}
\centering
\begin{scriptsize}
\begin{tabular}{cc}
\toprule
\multicolumn{2}{c}{CIFAR10} \\
\midrule
$\gamma$ & Accuracy \\
\midrule
0.1 &  $54.86 \; (\pm 3.16)$ \\
1 & 
$80.43 \; (\pm 1.15)$ \\
10 &  $79.85 \; (\pm 6.09)$ \\
\bottomrule
\end{tabular}
\quad
\begin{tabular}{cc}
\toprule
\multicolumn{2}{c}{CIFAR100} \\
\midrule
$\gamma$ & Accuracy \\
\midrule
0.1 & $11.22 \; (\pm 1.11)$ \\
1 &  $37.71 \; (\pm 3.68)$ \\
10 &  $38.20 \; (\pm 3.31)$ \\
\bottomrule
\end{tabular}
\end{scriptsize}
\medskip
\captionof{table}{Accuracies for different $\gamma$'s on CIFAR for the balanced scenario of \Cref{subsec:exp:est_bias}, with 2 times the standard deviation over 8 runs between parenthesis.}
\label{tab:table_accs_expe2}
\end{minipage}
\hfill
\begin{minipage}[c]{0.48\textwidth}
\centering
\begin{scriptsize}
\begin{tabular}{cc}
\toprule
\multicolumn{2}{c}{CIFAR10} \\
\midrule
$\gamma$ & Accuracy \\
\midrule
0.1 & $32.49 \; (\pm 8.71)$ \\
1 & $61.86 \; (\pm 6.21)$ \\
10 & $60.53 \; (\pm 4.88)$ \\
\bottomrule
\end{tabular}
\quad
\begin{tabular}{ccc}
\toprule
\multicolumn{2}{c}{CIFAR100} \\
\midrule
$\gamma$ & Accuracy\\
\midrule
0.1 & $6.40  \; (\pm 1.61)$ \\
1   & $19.35 \; (\pm 1.47)$ \\
10  & $19.95 \; (\pm 4.16)$ \\
\bottomrule 
\end{tabular}
\end{scriptsize}
\medskip
\captionof{table}{Accuracies for different $\gamma$'s on CIFAR for the imbalanced databases scenario of \Cref{subsec:exp:est_bias}, with 2 times the standard deviation over 8 runs.}
\label{tab:results_imbalanced_db_scenario}
\end{minipage}
\end{figure}

\par{\bf Imbalanced datasets scenario.}
Dividing the data over HSV bins can split CIFAR into
rather balanced bins, see \Cref{fig:bins_cifar10}.
However, in many practical scenarios, the different sources of data do not provide the same number of observations.
In that case, concatenating the databases can lead to very bad results.
We propose to study the relative performance of our debiaising methodology against the concatenation of the databases, for a long-tail distribution of the number of elements per dataset, as illustrated in \Cref{fig:bins_cifar10}.
Precisely, we fix $n_k \, \propto \, \alpha^k$ where $\alpha = 0.75$ and $n = \sum_{k=1}^K n_k = 50,000$.
In this experiment, we take into account the size of the datasets relatively to the size of the bins in the testing data using an additional multiplicative term in $\hat{\omega}_k$.
Results are provided in \Cref{tab:results_imbalanced_db_scenario}, highlighting the great impact of overlapping.

% \begin{table}[t]
%     \centering

%     \vspace{-0.2cm}
% \end{table}

\section{Related Works}\label{sec:related_work}
Learning visual recognition models in the presence of bias has recently been the subject of a good deal of attention in the computer vision literature.
For example, \cite{HendricksBSDR18} propose to correct for gender bias in captioning models by using masks that hide the gender information, while \cite{Kim_2019_CVPR} learn to dissociate an objective classification task to some associated counfounding bias using a known prior on the bias.
We hereby review some of the relevant works in this line of research.
\medskip

\par{\bf Bias and Transfer Learning.}
Since we consider bias as any significant difference between the training and testing distributions, learning with biased datasets is related to \emph{transfer learning}.
We recall that $\X$ (resp. $\Y$) denotes the input (resp. output) space associated to a computer vision task.
\cite{ASurveyTransferLearning} define \emph{domain adaptation} as the specific transfer learning setting
where both the task --- e.g., object detection or visual recognition --- and the input space $\X$ are the same for both training and testing data, but the train and test distribution are different, \textit{i.e.}, $p_{\text{train}}(x,y) \ne p_{\text{test}}(x, y)$ for some $(x, y) \in \X\times\Y$.
We denote by $z \in \mathcal{Z} = \X \times \Y$ any tuple $(x, y) \in \X\times\Y$.
Our work makes similar assumptions as domain adaptation,
but considers several training sets.
\medskip

\par{\bf Transferable Features.}
\cite{LongC0J15,GaninJMLR} address domain adaptation by learning a feature extractor $g$ that is invariant with respect to the shift between the domains, which makes the approach well suited to problems where only the marginal distribution $p(x)$ over $\mathcal{X}$ is different between training and testing.
Under this assumption, the posterior distributions satisfy $p_{\text{train}} (y \mid g(x)) = p_{\text{test}}(y \mid g(x))$ for any $(x, y) \in  \X\times\Y$, and we have $p_{\text{train}} (x) \ne p_{\text{test}}(x)$ for some $x \in \X$.
As a consequence, their approach is adapted to the case where the difference is reduced to that of the ``low-level'' visual appearance of the images (see \Cref{fig:three_types_bias}).
In contrast, \cite{ParametrizationMultiDomainCVPR} considers the domains to be types of images (e.g., medical, satellite), which makes learning invariant deep features undesirable.
In the facial recognition example, learning deep invariant features would discard information that is relevant for identification.
For those reasons, \cite{ParametrizationMultiDomainCVPR}, as
well as \cite{GuoSKGRF19}, considers all networks layers as possible candidates to adaptation.
Taking a different approach, \cite{Tzeng2015,LongZ0J16,Zhu2020} all explicitly model different train and test posterior probabilities $p(y \mid x)$ for the output, for example through specialized classifiers for both the target and source dataset \citep{Zhu2020}.
Most of the work on deep transfer focuses on adapting feature representations, but another popular approach in transfer learning is to transfer knowledge from instances \citep{ASurveyTransferLearning}.
\medskip

\par{\bf Instance Selection.}
Recent computer vision papers have proposed to select specific observations from a large scale training dataset to improve performance on the testing data.
For instance, \cite{CuiSSHB18} selects observations that are close to the target data, as measured by a generic image encoder, while \cite{GeY17} uses a similarity based
on low-level characteristics.
More similar to our work, \cite{Ngiam2018,Zhu2020} use importance weights for training instances, that are computed from a small set of observations that follow the target distribution.
Precisely, \cite{Ngiam2018} transfers information from the predicted distribution over labels of the adapted network, while \cite{Zhu2020} learns with a reinforcement learning selection policy guided by a performance metric on a validation set.
The objective of the importance function is to reweight any training instance $(x, y)$ by a quantity proportional to the likelihood ratio $\Phi(x, y) := p_{\text{test}}(x, y)/p_{\text{train}}(x, y)$.
While most papers assume the availability of a sample that follows $p_{\text{test}}$ to approximate the likelihood ratio $\Phi$, recent work in statistical machine learning have proposed to introduce milder assumptions on the relationship between $p_{\text{train}}$ and $p_{\text{test}}$ \citep{vogel2020weighted,laforgue2019statistical}.
These assumptions however typically require the machine learning practitioner to only have some \textit{a priori} knowledge on the testing data, which happens naturally in many situations, notably in industrial benchmarks.
For a facial recognition system, the providers often have accurate knowledge of some characteristics (e.g., nationality) of the target distribution but no access to the target data.
\cite{vogel2020weighted} assumed that a single training sample covers the whole testing distribution (with a dominated measure assumption), but \cite{laforgue2019statistical} relaxes that assumption by considering several source datasets.
\medskip

\par{\bf Multiple Source Datasets.}
%
% Proposition PL 15/07/21
%
Learning from several datasets has been studied in various contexts, such as domain adaptation \citep{MansourMR08,Ben-DavidBCKPV10}, or domain generalization \citep{LiYSH18}.
In the latter work, the authors optimize for all sources simultaneously, and find an interesting trade-off between the different competing objectives.
Our approach is closer to the one developed in \cite{laforgue2019statistical}, which consists in appropriately reweighting the observations generated by some biased sources.
Under several technical assumptions, the authors have extended the results of \cite{vardi1985} and have shown that the unbiased estimate of the target distribution thus obtained can be used to learn reliable decision rules.
Note however that an important limitation of these works is that the biasing functions are assumed to be exactly known.
In the present work, we learn a deep neural network for visual recognition with sampling weights derived from \cite{laforgue2019statistical}, but using approximate (learned) expressions for the biaising functions.
Note that \cite{Kim_2019_CVPR} also assumes the knowledge of bias information to adjust a neural network.
However, their approach is pretty incomparable to ours, insofar as it consists in penalizing the mutual information between the bias and the objective.

\section{Conclusion}\label{sec:conclusion}
In this paper, we build on the work by \cite{laforgue2019statistical} to learn reliable visual recognition decision functions from several biased datasets with approximated-on-the-fly biasing functions $\omega_k$.
However, this approach is not readily applicable to image data, as image datasets typically do not occupy the same portions of the space of all possible images.
To circumvent this problem, we propose to express the biasing functions either 1) by using additional information on the images, or 2) on smaller spaces by transforming the information contained in the images.
While our work demonstrates the effectiveness of a general methodology to learn with biased data, the choice of the biasing functions (or of the family to approximate them) should be performed on a case-by-case basis, and we cannot provide precise information on the expected effects of reweighting the data in full generality.
For these reasons, we consider the two most promising directions for future work to consist in: 1) finding a general methodology for choosing relevant approximations
to biaising functions, and 2) predicting the impact of the reweighting procedure before training, which is especially important when dealing with large models.

% \section*{Funding}
% An unnumbered section, e.g.\ \verb"\section*{Funding}", may be used for grant details, etc.\ if required and included \emph{in the non-anonymous version} before any Notes or References.

\newpage
\bibliographystyle{apacite}
\bibliography{library}

\begin{thebibliography}{}

\bibitem [\protect \citeauthoryear {%
Ben{-}David%
\ \protect \BOthers {.}}{%
Ben{-}David%
\ \protect \BOthers {.}}{%
{\protect \APACyear {2010}}%
}]{%
Ben-DavidBCKPV10}
\APACinsertmetastar {%
Ben-DavidBCKPV10}%
\begin{APACrefauthors}%
Ben{-}David, S.%
, Blitzer, J.%
, Crammer, K.%
, Kulesza, A.%
, Pereira, F.%
\BCBL {}\ \BBA {} Vaughan, J\BPBI W.%
\end{APACrefauthors}%
\unskip\
\newblock
\APACrefYearMonthDay{2010}{}{}.
\newblock
{\BBOQ}\APACrefatitle {A theory of learning from different domains} {A theory
  of learning from different domains}.{\BBCQ}
\newblock
\APACjournalVolNumPages{Machine Learning}{79}{1-2}{151--175}.
\PrintBackRefs{\CurrentBib}

\bibitem [\protect \citeauthoryear {%
Bousmalis%
, Trigeorgis%
, Silberman%
, Krishnan%
\BCBL {}\ \BBA {} Erhan%
}{%
Bousmalis%
\ \protect \BOthers {.}}{%
{\protect \APACyear {2016}}%
}]{%
BousmalisTSKE16}
\APACinsertmetastar {%
BousmalisTSKE16}%
\begin{APACrefauthors}%
Bousmalis, K.%
, Trigeorgis, G.%
, Silberman, N.%
, Krishnan, D.%
\BCBL {}\ \BBA {} Erhan, D.%
\end{APACrefauthors}%
\unskip\
\newblock
\APACrefYearMonthDay{2016}{}{}.
\newblock
{\BBOQ}\APACrefatitle {Domain Separation Networks} {Domain separation
  networks}.{\BBCQ}
\newblock
\BIn{} \APACrefbtitle {Advances in Neural Information Processing Systems 29:
  Annual Conference on Neural Information Processing Systems 2016} {Advances in
  neural information processing systems 29: Annual conference on neural
  information processing systems 2016}\ (\BPGS\ 343--351).
\PrintBackRefs{\CurrentBib}

\bibitem [\protect \citeauthoryear {%
Cui%
, Song%
, Sun%
, Howard%
\BCBL {}\ \BBA {} Belongie%
}{%
Cui%
\ \protect \BOthers {.}}{%
{\protect \APACyear {2018}}%
}]{%
CuiSSHB18}
\APACinsertmetastar {%
CuiSSHB18}%
\begin{APACrefauthors}%
Cui, Y.%
, Song, Y.%
, Sun, C.%
, Howard, A.%
\BCBL {}\ \BBA {} Belongie, S\BPBI J.%
\end{APACrefauthors}%
\unskip\
\newblock
\APACrefYearMonthDay{2018}{}{}.
\newblock
{\BBOQ}\APACrefatitle {Large Scale Fine-Grained Categorization and
  Domain-Specific Transfer Learning} {Large scale fine-grained categorization
  and domain-specific transfer learning}.{\BBCQ}
\newblock
\BIn{} \APACrefbtitle {2018 {IEEE} Conference on Computer Vision and Pattern
  Recognition, {CVPR} 2018} {2018 {IEEE} conference on computer vision and
  pattern recognition, {CVPR} 2018}\ (\BPGS\ 4109--4118).
\newblock
\APACaddressPublisher{}{{IEEE} Computer Society}.
\PrintBackRefs{\CurrentBib}

\bibitem [\protect \citeauthoryear {%
Devroye%
, Györfi%
\BCBL {}\ \BBA {} Lugosi%
}{%
Devroye%
\ \protect \BOthers {.}}{%
{\protect \APACyear {1996}}%
}]{%
Devroye1996}
\APACinsertmetastar {%
Devroye1996}%
\begin{APACrefauthors}%
Devroye, L.%
, Györfi, L.%
\BCBL {}\ \BBA {} Lugosi, G.%
\end{APACrefauthors}%
\unskip\
\newblock
\APACrefYear{1996}.
\newblock
\APACrefbtitle {A Probabilistic Theory of Pattern Recognition} {A probabilistic
  theory of pattern recognition}.
\newblock
\APACaddressPublisher{}{Springer}.
\PrintBackRefs{\CurrentBib}

\bibitem [\protect \citeauthoryear {%
Ganin%
\ \protect \BOthers {.}}{%
Ganin%
\ \protect \BOthers {.}}{%
{\protect \APACyear {2016}}%
}]{%
GaninJMLR}
\APACinsertmetastar {%
GaninJMLR}%
\begin{APACrefauthors}%
Ganin, Y.%
, Ustinova, E.%
, Ajakan, H.%
, Germain, P.%
, Larochelle, H.%
, Laviolette, F.%
\BDBL {}Lempitsky, V.%
\end{APACrefauthors}%
\unskip\
\newblock
\APACrefYearMonthDay{2016}{}{}.
\newblock
{\BBOQ}\APACrefatitle {Domain-Adversarial Training of Neural Networks}
  {Domain-adversarial training of neural networks}.{\BBCQ}
\newblock
\APACjournalVolNumPages{Journal of Machine Learning Research}{}{}{}.
\PrintBackRefs{\CurrentBib}

\bibitem [\protect \citeauthoryear {%
Ge%
\ \BBA {} Yu%
}{%
Ge%
\ \BBA {} Yu%
}{%
{\protect \APACyear {2017}}%
}]{%
GeY17}
\APACinsertmetastar {%
GeY17}%
\begin{APACrefauthors}%
Ge, W.%
\BCBT {}\ \BBA {} Yu, Y.%
\end{APACrefauthors}%
\unskip\
\newblock
\APACrefYearMonthDay{2017}{}{}.
\newblock
{\BBOQ}\APACrefatitle {Borrowing Treasures from the Wealthy: Deep Transfer
  Learning through Selective Joint Fine-Tuning} {Borrowing treasures from the
  wealthy: Deep transfer learning through selective joint fine-tuning}.{\BBCQ}
\newblock
\BIn{} \APACrefbtitle {2017 {IEEE} Conference on Computer Vision and Pattern
  Recognition, {CVPR} 2017} {2017 {IEEE} conference on computer vision and
  pattern recognition, {CVPR} 2017}\ (\BPGS\ 10--19).
\newblock
\APACaddressPublisher{}{{IEEE} Computer Society}.
\PrintBackRefs{\CurrentBib}

\bibitem [\protect \citeauthoryear {%
Gill%
, Vardi%
\BCBL {}\ \BBA {} Wellner%
}{%
Gill%
\ \protect \BOthers {.}}{%
{\protect \APACyear {1988}}%
}]{%
gill1988}
\APACinsertmetastar {%
gill1988}%
\begin{APACrefauthors}%
Gill, R\BPBI D.%
, Vardi, Y.%
\BCBL {}\ \BBA {} Wellner, J\BPBI A.%
\end{APACrefauthors}%
\unskip\
\newblock
\APACrefYearMonthDay{1988}{}{}.
\newblock
{\BBOQ}\APACrefatitle {Large Sample Theory of Empirical Distributions in Biased
  Sampling Models} {Large sample theory of empirical distributions in biased
  sampling models}.{\BBCQ}
\newblock
\APACjournalVolNumPages{Annals of Statistics}{16}{3}{1069--1112}.
\PrintBackRefs{\CurrentBib}

\bibitem [\protect \citeauthoryear {%
Grother%
\ \BBA {} Ngan%
}{%
Grother%
\ \BBA {} Ngan%
}{%
{\protect \APACyear {2019}}%
}]{%
Grother2019}
\APACinsertmetastar {%
Grother2019}%
\begin{APACrefauthors}%
Grother, P.%
\BCBT {}\ \BBA {} Ngan, M.%
\end{APACrefauthors}%
\unskip\
\newblock
\APACrefYear{2019}.
\newblock
\APACrefbtitle {{Face Recognition Vendor Test (FRVT) --- Performance of
  Automated Gender Classification Algorithms}} {{Face Recognition Vendor Test
  (FRVT) --- Performance of Automated Gender Classification Algorithms}}\
  (\BNUM\ NISTIR 8052).
\PrintBackRefs{\CurrentBib}

\bibitem [\protect \citeauthoryear {%
Guo%
\ \protect \BOthers {.}}{%
Guo%
\ \protect \BOthers {.}}{%
{\protect \APACyear {2019}}%
}]{%
GuoSKGRF19}
\APACinsertmetastar {%
GuoSKGRF19}%
\begin{APACrefauthors}%
Guo, Y.%
, Shi, H.%
, Kumar, A.%
, Grauman, K.%
, Rosing, T.%
\BCBL {}\ \BBA {} Feris, R\BPBI S.%
\end{APACrefauthors}%
\unskip\
\newblock
\APACrefYearMonthDay{2019}{}{}.
\newblock
{\BBOQ}\APACrefatitle {SpotTune: Transfer Learning Through Adaptive
  Fine-Tuning} {Spottune: Transfer learning through adaptive
  fine-tuning}.{\BBCQ}
\newblock
\BIn{} \APACrefbtitle {{IEEE} Conference on Computer Vision and Pattern
  Recognition, {CVPR} 2019, Long Beach, CA, USA, June 16-20, 2019} {{IEEE}
  conference on computer vision and pattern recognition, {CVPR} 2019, long
  beach, ca, usa, june 16-20, 2019}\ (\BPGS\ 4805--4814).
\newblock
\APACaddressPublisher{}{Computer Vision Foundation / {IEEE}}.
\PrintBackRefs{\CurrentBib}

\bibitem [\protect \citeauthoryear {%
Guo%
, Zhang%
, Hu%
, He%
\BCBL {}\ \BBA {} Gao%
}{%
Guo%
\ \protect \BOthers {.}}{%
{\protect \APACyear {2016}}%
}]{%
y2016msceleb1m}
\APACinsertmetastar {%
y2016msceleb1m}%
\begin{APACrefauthors}%
Guo, Y.%
, Zhang, L.%
, Hu, Y.%
, He, X.%
\BCBL {}\ \BBA {} Gao, J.%
\end{APACrefauthors}%
\unskip\
\newblock
\APACrefYearMonthDay{2016}{}{}.
\newblock
{\BBOQ}\APACrefatitle {MS-Celeb-1M: {A} Dataset and Benchmark for Large-Scale
  Face Recognition} {Ms-celeb-1m: {A} dataset and benchmark for large-scale
  face recognition}.{\BBCQ}
\newblock
\BIn{} \APACrefbtitle {Computer Vision - {ECCV} 2016 - 14th European
  Conference, Proceedings, Part {III}} {Computer vision - {ECCV} 2016 - 14th
  european conference, proceedings, part {III}}\ (\BVOL\ 9907, \BPGS\ 87--102).
\newblock
\APACaddressPublisher{}{Springer}.
\PrintBackRefs{\CurrentBib}

\bibitem [\protect \citeauthoryear {%
He%
, Zhang%
, Ren%
\BCBL {}\ \BBA {} Sun%
}{%
He%
\ \protect \BOthers {.}}{%
{\protect \APACyear {2015}}%
}]{%
He2015}
\APACinsertmetastar {%
He2015}%
\begin{APACrefauthors}%
He, K.%
, Zhang, X.%
, Ren, S.%
\BCBL {}\ \BBA {} Sun, J.%
\end{APACrefauthors}%
\unskip\
\newblock
\APACrefYearMonthDay{2015}{}{}.
\newblock
{\BBOQ}\APACrefatitle {Deep Residual Learning for Image Recognition} {Deep
  residual learning for image recognition}.{\BBCQ}
\newblock
\APACjournalVolNumPages{CoRR}{abs/1512.03385}{}{}.
\PrintBackRefs{\CurrentBib}

\bibitem [\protect \citeauthoryear {%
Hendricks%
, Burns%
, Saenko%
, Darrell%
\BCBL {}\ \BBA {} Rohrbach%
}{%
Hendricks%
\ \protect \BOthers {.}}{%
{\protect \APACyear {2018}}%
}]{%
HendricksBSDR18}
\APACinsertmetastar {%
HendricksBSDR18}%
\begin{APACrefauthors}%
Hendricks, L\BPBI A.%
, Burns, K.%
, Saenko, K.%
, Darrell, T.%
\BCBL {}\ \BBA {} Rohrbach, A.%
\end{APACrefauthors}%
\unskip\
\newblock
\APACrefYearMonthDay{2018}{}{}.
\newblock
{\BBOQ}\APACrefatitle {Women Also Snowboard: Overcoming Bias in Captioning
  Models} {Women also snowboard: Overcoming bias in captioning models}.{\BBCQ}
\newblock
\BIn{} \APACrefbtitle {Computer Vision - {ECCV} 2018 - 15th European
  Conference} {Computer vision - {ECCV} 2018 - 15th european conference}\
  (\BVOL\ 11207, \BPGS\ 793--811).
\newblock
\APACaddressPublisher{}{Springer}.
\PrintBackRefs{\CurrentBib}

\bibitem [\protect \citeauthoryear {%
Huang%
, Ramesh%
, Berg%
\BCBL {}\ \BBA {} Learned-Miller%
}{%
Huang%
\ \protect \BOthers {.}}{%
{\protect \APACyear {2007}}%
}]{%
LFWTech}
\APACinsertmetastar {%
LFWTech}%
\begin{APACrefauthors}%
Huang, G\BPBI B.%
, Ramesh, M.%
, Berg, T.%
\BCBL {}\ \BBA {} Learned-Miller, E.%
\end{APACrefauthors}%
\unskip\
\newblock
\APACrefYear{2007}.
\newblock
\APACrefbtitle {Labeled Faces in the Wild: A Database for Studying Face
  Recognition in Unconstrained Environments} {Labeled faces in the wild: A
  database for studying face recognition in unconstrained environments}\
  (\BNUM\ 07-49).
\PrintBackRefs{\CurrentBib}

\bibitem [\protect \citeauthoryear {%
Kim%
, Kim%
, Kim%
, Kim%
\BCBL {}\ \BBA {} Kim%
}{%
Kim%
\ \protect \BOthers {.}}{%
{\protect \APACyear {2019}}%
}]{%
Kim_2019_CVPR}
\APACinsertmetastar {%
Kim_2019_CVPR}%
\begin{APACrefauthors}%
Kim, B.%
, Kim, H.%
, Kim, K.%
, Kim, S.%
\BCBL {}\ \BBA {} Kim, J.%
\end{APACrefauthors}%
\unskip\
\newblock
\APACrefYearMonthDay{2019}{June}{}.
\newblock
{\BBOQ}\APACrefatitle {Learning Not to Learn: Training Deep Neural Networks
  With Biased Data} {Learning not to learn: Training deep neural networks with
  biased data}.{\BBCQ}
\newblock
\BIn{} \APACrefbtitle {Proceedings of the IEEE/CVF Conference on Computer
  Vision and Pattern Recognition (CVPR).} {Proceedings of the ieee/cvf
  conference on computer vision and pattern recognition (cvpr).}
\PrintBackRefs{\CurrentBib}

\bibitem [\protect \citeauthoryear {%
Laforgue%
\ \BBA {} Cl{\'{e}}men{\c{c}}on%
}{%
Laforgue%
\ \BBA {} Cl{\'{e}}men{\c{c}}on%
}{%
{\protect \APACyear {2019}}%
}]{%
laforgue2019statistical}
\APACinsertmetastar {%
laforgue2019statistical}%
\begin{APACrefauthors}%
Laforgue, P.%
\BCBT {}\ \BBA {} Cl{\'{e}}men{\c{c}}on, S.%
\end{APACrefauthors}%
\unskip\
\newblock
\APACrefYearMonthDay{2019}{}{}.
\newblock
{\BBOQ}\APACrefatitle {Statistical Learning from Biased Training Samples}
  {Statistical learning from biased training samples}.{\BBCQ}
\newblock
\APACjournalVolNumPages{CoRR}{abs/1906.12304}{}{}.
\PrintBackRefs{\CurrentBib}

\bibitem [\protect \citeauthoryear {%
Li%
, Yang%
, Song%
\BCBL {}\ \BBA {} Hospedales%
}{%
Li%
\ \protect \BOthers {.}}{%
{\protect \APACyear {2018}}%
}]{%
LiYSH18}
\APACinsertmetastar {%
LiYSH18}%
\begin{APACrefauthors}%
Li, D.%
, Yang, Y.%
, Song, Y.%
\BCBL {}\ \BBA {} Hospedales, T\BPBI M.%
\end{APACrefauthors}%
\unskip\
\newblock
\APACrefYearMonthDay{2018}{}{}.
\newblock
{\BBOQ}\APACrefatitle {Learning to Generalize: Meta-Learning for Domain
  Generalization} {Learning to generalize: Meta-learning for domain
  generalization}.{\BBCQ}
\newblock
\BIn{} \APACrefbtitle {Proceedings of the Thirty-Second {AAAI} Conference on
  Artificial Intelligence, the 30th innovative Applications of Artificial
  Intelligence, and the 8th {AAAI} Symposium on Educational Advances in
  Artificial Intelligence} {Proceedings of the thirty-second {AAAI} conference
  on artificial intelligence, the 30th innovative applications of artificial
  intelligence, and the 8th {AAAI} symposium on educational advances in
  artificial intelligence}\ (\BPGS\ 3490--3497).
\newblock
\APACaddressPublisher{}{{AAAI} Press}.
\PrintBackRefs{\CurrentBib}

\bibitem [\protect \citeauthoryear {%
Long%
, Cao%
, Wang%
\BCBL {}\ \BBA {} Jordan%
}{%
Long%
\ \protect \BOthers {.}}{%
{\protect \APACyear {2015}}%
}]{%
LongC0J15}
\APACinsertmetastar {%
LongC0J15}%
\begin{APACrefauthors}%
Long, M.%
, Cao, Y.%
, Wang, J.%
\BCBL {}\ \BBA {} Jordan, M\BPBI I.%
\end{APACrefauthors}%
\unskip\
\newblock
\APACrefYearMonthDay{2015}{}{}.
\newblock
{\BBOQ}\APACrefatitle {Learning Transferable Features with Deep Adaptation
  Networks} {Learning transferable features with deep adaptation
  networks}.{\BBCQ}
\newblock
\BIn{} \APACrefbtitle {Proceedings of the 32nd International Conference on
  Machine Learning, {ICML} 2015} {Proceedings of the 32nd international
  conference on machine learning, {ICML} 2015}\ (\BVOL~37, \BPGS\ 97--105).
\newblock
\APACaddressPublisher{}{JMLR.org}.
\PrintBackRefs{\CurrentBib}

\bibitem [\protect \citeauthoryear {%
Long%
, Zhu%
, Wang%
\BCBL {}\ \BBA {} Jordan%
}{%
Long%
\ \protect \BOthers {.}}{%
{\protect \APACyear {2016}}%
}]{%
LongZ0J16}
\APACinsertmetastar {%
LongZ0J16}%
\begin{APACrefauthors}%
Long, M.%
, Zhu, H.%
, Wang, J.%
\BCBL {}\ \BBA {} Jordan, M\BPBI I.%
\end{APACrefauthors}%
\unskip\
\newblock
\APACrefYearMonthDay{2016}{}{}.
\newblock
{\BBOQ}\APACrefatitle {Unsupervised Domain Adaptation with Residual Transfer
  Networks} {Unsupervised domain adaptation with residual transfer
  networks}.{\BBCQ}
\newblock
\BIn{} \APACrefbtitle {Advances in Neural Information Processing Systems 29:
  Annual Conference on Neural Information Processing Systems 2016} {Advances in
  neural information processing systems 29: Annual conference on neural
  information processing systems 2016}\ (\BPGS\ 136--144).
\PrintBackRefs{\CurrentBib}

\bibitem [\protect \citeauthoryear {%
Mansour%
, Mohri%
\BCBL {}\ \BBA {} Rostamizadeh%
}{%
Mansour%
\ \protect \BOthers {.}}{%
{\protect \APACyear {2008}}%
}]{%
MansourMR08}
\APACinsertmetastar {%
MansourMR08}%
\begin{APACrefauthors}%
Mansour, Y.%
, Mohri, M.%
\BCBL {}\ \BBA {} Rostamizadeh, A.%
\end{APACrefauthors}%
\unskip\
\newblock
\APACrefYearMonthDay{2008}{}{}.
\newblock
{\BBOQ}\APACrefatitle {Domain Adaptation with Multiple Sources} {Domain
  adaptation with multiple sources}.{\BBCQ}
\newblock
\BIn{} \APACrefbtitle {Advances in Neural Information Processing Systems 21}
  {Advances in neural information processing systems 21}\ (\BPGS\ 1041--1048).
\newblock
\APACaddressPublisher{}{Curran Associates, Inc.}
\PrintBackRefs{\CurrentBib}

\bibitem [\protect \citeauthoryear {%
Menon%
\ \protect \BOthers {.}}{%
Menon%
\ \protect \BOthers {.}}{%
{\protect \APACyear {2021}}%
}]{%
MenonJRJVK21}
\APACinsertmetastar {%
MenonJRJVK21}%
\begin{APACrefauthors}%
Menon, A\BPBI K.%
, Jayasumana, S.%
, Rawat, A\BPBI S.%
, Jain, H.%
, Veit, A.%
\BCBL {}\ \BBA {} Kumar, S.%
\end{APACrefauthors}%
\unskip\
\newblock
\APACrefYearMonthDay{2021}{}{}.
\newblock
{\BBOQ}\APACrefatitle {Long-tail learning via logit adjustment} {Long-tail
  learning via logit adjustment}.{\BBCQ}
\newblock
\BIn{} \APACrefbtitle {9th International Conference on Learning
  Representations, {ICLR} 2021.} {9th international conference on learning
  representations, {ICLR} 2021.}
\newblock
\APACaddressPublisher{}{OpenReview.net}.
\PrintBackRefs{\CurrentBib}

\bibitem [\protect \citeauthoryear {%
Ngiam%
\ \protect \BOthers {.}}{%
Ngiam%
\ \protect \BOthers {.}}{%
{\protect \APACyear {2018}}%
}]{%
Ngiam2018}
\APACinsertmetastar {%
Ngiam2018}%
\begin{APACrefauthors}%
Ngiam, J.%
, Peng, D.%
, Vasudevan, V.%
, Kornblith, S.%
, Le, Q\BPBI V.%
\BCBL {}\ \BBA {} Pang, R.%
\end{APACrefauthors}%
\unskip\
\newblock
\APACrefYearMonthDay{2018}{}{}.
\newblock
{\BBOQ}\APACrefatitle {Domain Adaptive Transfer Learning with Specialist
  Models} {Domain adaptive transfer learning with specialist models}.{\BBCQ}
\newblock
\APACjournalVolNumPages{CoRR}{abs/1811.07056}{}{}.
\PrintBackRefs{\CurrentBib}

\bibitem [\protect \citeauthoryear {%
{Pan}%
\ \BBA {} {Yang}%
}{%
{Pan}%
\ \BBA {} {Yang}%
}{%
{\protect \APACyear {2010}}%
}]{%
ASurveyTransferLearning}
\APACinsertmetastar {%
ASurveyTransferLearning}%
\begin{APACrefauthors}%
{Pan}, S\BPBI J.%
\BCBT {}\ \BBA {} {Yang}, Q.%
\end{APACrefauthors}%
\unskip\
\newblock
\APACrefYearMonthDay{2010}{}{}.
\newblock
{\BBOQ}\APACrefatitle {A Survey on Transfer Learning} {A survey on transfer
  learning}.{\BBCQ}
\newblock
\APACjournalVolNumPages{IEEE Transactions on Knowledge and Data
  Engineering}{22}{10}{1345-1359}.
\PrintBackRefs{\CurrentBib}

\bibitem [\protect \citeauthoryear {%
Paszke%
\ \protect \BOthers {.}}{%
Paszke%
\ \protect \BOthers {.}}{%
{\protect \APACyear {2019}}%
}]{%
pytorch}
\APACinsertmetastar {%
pytorch}%
\begin{APACrefauthors}%
Paszke, A.%
, Gross, S.%
, Massa, F.%
, Lerer, A.%
, Bradbury, J.%
, Chanan, G.%
\BDBL {}Chintala, S.%
\end{APACrefauthors}%
\unskip\
\newblock
\APACrefYearMonthDay{2019}{}{}.
\newblock
{\BBOQ}\APACrefatitle {PyTorch: An Imperative Style, High-Performance Deep
  Learning Library} {Pytorch: An imperative style, high-performance deep
  learning library}.{\BBCQ}
\newblock
\BIn{} \APACrefbtitle {Advances in Neural Information Processing Systems 32}
  {Advances in neural information processing systems 32}\ (\BPGS\ 8024--8035).
\newblock
\APACaddressPublisher{}{Curran Associates, Inc.}
\PrintBackRefs{\CurrentBib}

\bibitem [\protect \citeauthoryear {%
Phillips%
, Jiang%
, Narvekar%
, Ayyad%
\BCBL {}\ \BBA {} O'Toole%
}{%
Phillips%
\ \protect \BOthers {.}}{%
{\protect \APACyear {2011}}%
}]{%
OtherFaceRecognition2}
\APACinsertmetastar {%
OtherFaceRecognition2}%
\begin{APACrefauthors}%
Phillips, P\BPBI J.%
, Jiang, F.%
, Narvekar, A.%
, Ayyad, J\BPBI H.%
\BCBL {}\ \BBA {} O'Toole, A\BPBI J.%
\end{APACrefauthors}%
\unskip\
\newblock
\APACrefYearMonthDay{2011}{}{}.
\newblock
{\BBOQ}\APACrefatitle {An other-race effect for face recognition algorithms}
  {An other-race effect for face recognition algorithms}.{\BBCQ}
\newblock
\APACjournalVolNumPages{{ACM} Transactions on Applied
  Perception}{8}{2}{14:1--14:11}.
\PrintBackRefs{\CurrentBib}

\bibitem [\protect \citeauthoryear {%
Quionero-Candela%
, Sugiyama%
, Schwaighofer%
\BCBL {}\ \BBA {} Lawrence%
}{%
Quionero-Candela%
\ \protect \BOthers {.}}{%
{\protect \APACyear {2009}}%
}]{%
quionero2009dataset}
\APACinsertmetastar {%
quionero2009dataset}%
\begin{APACrefauthors}%
Quionero-Candela, J.%
, Sugiyama, M.%
, Schwaighofer, A.%
\BCBL {}\ \BBA {} Lawrence, N.%
\end{APACrefauthors}%
\unskip\
\newblock
\APACrefYear{2009}.
\newblock
\APACrefbtitle {Dataset shift in machine learning} {Dataset shift in machine
  learning}.
\newblock
\APACaddressPublisher{}{The MIT Press}.
\PrintBackRefs{\CurrentBib}

\bibitem [\protect \citeauthoryear {%
{Rebuffi}%
, {Vedaldi}%
\BCBL {}\ \BBA {} {Bilen}%
}{%
{Rebuffi}%
\ \protect \BOthers {.}}{%
{\protect \APACyear {2018}}%
}]{%
ParametrizationMultiDomainCVPR}
\APACinsertmetastar {%
ParametrizationMultiDomainCVPR}%
\begin{APACrefauthors}%
{Rebuffi}, S.%
, {Vedaldi}, A.%
\BCBL {}\ \BBA {} {Bilen}, H.%
\end{APACrefauthors}%
\unskip\
\newblock
\APACrefYearMonthDay{2018}{}{}.
\newblock
{\BBOQ}\APACrefatitle {Efficient Parametrization of Multi-domain Deep Neural
  Networks} {Efficient parametrization of multi-domain deep neural
  networks}.{\BBCQ}
\newblock
\BIn{} \APACrefbtitle {2018 IEEE/CVF Conference on Computer Vision and Pattern
  Recognition} {2018 ieee/cvf conference on computer vision and pattern
  recognition}\ (\BPG~8119-8127).
\PrintBackRefs{\CurrentBib}

\bibitem [\protect \citeauthoryear {%
Sugiyama%
\ \BBA {} Kawanabe%
}{%
Sugiyama%
\ \BBA {} Kawanabe%
}{%
{\protect \APACyear {2012}}%
}]{%
sugiyama2012machine}
\APACinsertmetastar {%
sugiyama2012machine}%
\begin{APACrefauthors}%
Sugiyama, M.%
\BCBT {}\ \BBA {} Kawanabe, M.%
\end{APACrefauthors}%
\unskip\
\newblock
\APACrefYear{2012}.
\newblock
\APACrefbtitle {Machine learning in non-stationary environments: Introduction
  to covariate shift adaptation} {Machine learning in non-stationary
  environments: Introduction to covariate shift adaptation}.
\newblock
\APACaddressPublisher{}{The MIT Press}.
\PrintBackRefs{\CurrentBib}

\bibitem [\protect \citeauthoryear {%
Tzeng%
, Hoffman%
, Darrell%
\BCBL {}\ \BBA {} Saenko%
}{%
Tzeng%
\ \protect \BOthers {.}}{%
{\protect \APACyear {2015}}%
}]{%
Tzeng2015}
\APACinsertmetastar {%
Tzeng2015}%
\begin{APACrefauthors}%
Tzeng, E.%
, Hoffman, J.%
, Darrell, T.%
\BCBL {}\ \BBA {} Saenko, K.%
\end{APACrefauthors}%
\unskip\
\newblock
\APACrefYearMonthDay{2015}{}{}.
\newblock
{\BBOQ}\APACrefatitle {Simultaneous Deep Transfer Across Domains and Tasks}
  {Simultaneous deep transfer across domains and tasks}.{\BBCQ}
\newblock
\BIn{} \APACrefbtitle {Proceedings of the 2015 IEEE International Conference on
  Computer Vision (ICCV)} {Proceedings of the 2015 ieee international
  conference on computer vision (iccv)}\ (\BPG~4068–4076).
\newblock
\APACaddressPublisher{USA}{IEEE Computer Society}.
\PrintBackRefs{\CurrentBib}

\bibitem [\protect \citeauthoryear {%
van~der Vaart%
\ \BBA {} Wellner%
}{%
van~der Vaart%
\ \BBA {} Wellner%
}{%
{\protect \APACyear {1996}}%
}]{%
VanderVaart1996}
\APACinsertmetastar {%
VanderVaart1996}%
\begin{APACrefauthors}%
van~der Vaart, A.%
\BCBT {}\ \BBA {} Wellner, J.%
\end{APACrefauthors}%
\unskip\
\newblock
\APACrefYear{1996}.
\newblock
\APACrefbtitle {{Weak convergence and empirical processes}} {{Weak convergence
  and empirical processes}}.
\newblock
\APACaddressPublisher{}{Springer-Verlag}.
\PrintBackRefs{\CurrentBib}

\bibitem [\protect \citeauthoryear {%
Vardi%
}{%
Vardi%
}{%
{\protect \APACyear {1985}}%
}]{%
vardi1985}
\APACinsertmetastar {%
vardi1985}%
\begin{APACrefauthors}%
Vardi, Y.%
\end{APACrefauthors}%
\unskip\
\newblock
\APACrefYearMonthDay{1985}{03}{}.
\newblock
{\BBOQ}\APACrefatitle {Empirical Distributions in Selection Bias Models}
  {Empirical distributions in selection bias models}.{\BBCQ}
\newblock
\APACjournalVolNumPages{Ann. Statist.}{13}{1}{178--203}.
\newblock
\begin{APACrefDOI} \doi{10.1214/aos/1176346585} \end{APACrefDOI}
\PrintBackRefs{\CurrentBib}

\bibitem [\protect \citeauthoryear {%
Vinyals%
, Blundell%
, Lillicrap%
, Kavukcuoglu%
\BCBL {}\ \BBA {} Wierstra%
}{%
Vinyals%
\ \protect \BOthers {.}}{%
{\protect \APACyear {2016}}%
}]{%
VinyalsBLKW16}
\APACinsertmetastar {%
VinyalsBLKW16}%
\begin{APACrefauthors}%
Vinyals, O.%
, Blundell, C.%
, Lillicrap, T.%
, Kavukcuoglu, K.%
\BCBL {}\ \BBA {} Wierstra, D.%
\end{APACrefauthors}%
\unskip\
\newblock
\APACrefYearMonthDay{2016}{}{}.
\newblock
{\BBOQ}\APACrefatitle {Matching Networks for One Shot Learning} {Matching
  networks for one shot learning}.{\BBCQ}
\newblock
\BIn{} \APACrefbtitle {Advances in Neural Information Processing Systems 29:
  Annual Conference on Neural Information Processing Systems 2016} {Advances in
  neural information processing systems 29: Annual conference on neural
  information processing systems 2016}\ (\BPGS\ 3630--3638).
\PrintBackRefs{\CurrentBib}

\bibitem [\protect \citeauthoryear {%
Vogel%
, Achab%
, Cl{\'{e}}men{\c{c}}on%
\BCBL {}\ \BBA {} Tillier%
}{%
Vogel%
\ \protect \BOthers {.}}{%
{\protect \APACyear {2020}}%
}]{%
vogel2020weighted}
\APACinsertmetastar {%
vogel2020weighted}%
\begin{APACrefauthors}%
Vogel, R.%
, Achab, M.%
, Cl{\'{e}}men{\c{c}}on, S.%
\BCBL {}\ \BBA {} Tillier, C.%
\end{APACrefauthors}%
\unskip\
\newblock
\APACrefYearMonthDay{2020}{}{}.
\newblock
{\BBOQ}\APACrefatitle {Weighted Empirical Risk Minimization: Sample Selection
  Bias Correction based on Importance Sampling} {Weighted empirical risk
  minimization: Sample selection bias correction based on importance
  sampling}.{\BBCQ}
\newblock
\APACjournalVolNumPages{CoRR}{abs/2002.05145}{}{}.
\PrintBackRefs{\CurrentBib}

\bibitem [\protect \citeauthoryear {%
Wang%
, Deng%
, Hu%
, Tao%
\BCBL {}\ \BBA {} Huang%
}{%
Wang%
\ \protect \BOthers {.}}{%
{\protect \APACyear {2019}}%
}]{%
racialfaces}
\APACinsertmetastar {%
racialfaces}%
\begin{APACrefauthors}%
Wang, M.%
, Deng, W.%
, Hu, J.%
, Tao, X.%
\BCBL {}\ \BBA {} Huang, Y.%
\end{APACrefauthors}%
\unskip\
\newblock
\APACrefYearMonthDay{2019}{}{}.
\newblock
{\BBOQ}\APACrefatitle {Racial Faces in the Wild: Reducing Racial Bias by
  Information Maximization Adaptation Network} {Racial faces in the wild:
  Reducing racial bias by information maximization adaptation network}.{\BBCQ}
\newblock
\BIn{} \APACrefbtitle {2019 {IEEE/CVF} International Conference on Computer
  Vision, {ICCV} 2019} {2019 {IEEE/CVF} international conference on computer
  vision, {ICCV} 2019}\ (\BPGS\ 692--702).
\newblock
\APACaddressPublisher{}{{IEEE}}.
\PrintBackRefs{\CurrentBib}

\bibitem [\protect \citeauthoryear {%
Zhu%
, Arik%
, Yang%
\BCBL {}\ \BBA {} Pfister%
}{%
Zhu%
\ \protect \BOthers {.}}{%
{\protect \APACyear {2020}}%
}]{%
Zhu2020}
\APACinsertmetastar {%
Zhu2020}%
\begin{APACrefauthors}%
Zhu, L.%
, Arik, S.%
, Yang, Y.%
\BCBL {}\ \BBA {} Pfister, T.%
\end{APACrefauthors}%
\unskip\
\newblock
\APACrefYearMonthDay{2020}{}{}.
\newblock
{\BBOQ}\APACrefatitle {Learning to Transfer Learn: Reinforcement Learning-Based
  Selection for Adaptive Transfer Learning} {Learning to transfer learn:
  Reinforcement learning-based selection for adaptive transfer
  learning}.{\BBCQ}
\newblock
\APACjournalVolNumPages{arXiv preprint arXiv:1908.11406}{}{}{}.
\PrintBackRefs{\CurrentBib}

\end{thebibliography}

\appendix
\section{Proof of Theorem \ref{thm:approx}}
\label{apx:proof}

The proof of \Cref{thm:approx} follows that of Theorem~1 in \citet{laforgue2019statistical}.
For conciseness, the latter paper will be denoted LC19 in the rest of the proof.
Note that for simplicity we consider here datasets with fixed proportions $\lambda_k = n_k/n$, such that Assumption~4 in LC19 is by definition satisfied with $C_\lambda = 0$.
The extension to random proportions (satisfying Assumption~4 in LC19)  is immediate and left to the reader.
We proceed by adapting each of the intermediary results of LC19.
\medskip

\par{\bf Adapting Proposition 1 from LC19.}
Proposition 1 in LC19 builds on the fact that the graph $\widehat{G}_n$ defined in equation (3.2) therein is strongly connected with high probability if \Cref{hyp:connect} holds true.
Since we use the approximate biasing functions, the directed graph we should check the strong connectivity of is defined as $\widehat{G}_n^\mathrm{approx}$, with vertices in $\{1, \ldots, K\}$, and edge $k \rightarrow l$ if and only if
\[
\int \mathbb{I}\{\hat{\omega}_k(z) > 0\}d\hat{p}_l(z) > 0\,.
\]
Now, assume that $n \ge 4C_\omega^2/\varepsilon^2$.
For any $z \in \mathcal{Z}$ and $k \in \{1, \ldots, K\}$, we have
\begin{equation}\label{eq:lower_omega}
\hat{\omega}_k(z) \ge \omega_k(z) - \frac{C_\omega}{\sqrt{n}} \ge \varepsilon\cdot\mathbb{I}\{\omega_k(z) > 0\} - \frac{\varepsilon}{2} \ge \frac{\varepsilon}{2}\cdot\mathbb{I}\{\omega_k(z) > 0\}\,,
\end{equation}
where we have also used \Cref{hyp:bounded_omega}.
Hence, we have $\mathbb{I}\{\hat{\omega}_k(z) > 0\} \ge \mathbb{I}\{\omega_k(z) > 0\}$, and any edge in $\widehat{G}_n$ will also be an edge in $\widehat{G}_n^\mathrm{approx}$.
Thus $\widehat{G}_n^\mathrm{approx}$ is strongly connected with at least the same probability as $\widehat{G}_n$, and we have established an analog of the first claim of Proposition~1 in LC19 holds (whenever $n \ge 4C_\omega^2/\varepsilon^2$).

Regarding the second claim, we can follow the path from LC19, the only difference being that we use $\hat{\omega}_k(z) \ge  (\varepsilon/2)\mathbb{I}\{\omega_k(z) > 0\}$, that we have established in \eqref{eq:lower_omega}, rather than $\omega_k(z) \ge  \varepsilon\,\mathbb{I}\{\omega_k(z) > 0\}$.
Note that some constants may change, but still depend exclusively on the parameters of the problem, such that we omit their explicit formula for simplicity.
\medskip

\par{\bf Adapting Proposition 4 from LC19.}
Using the approximate bias functions $\hat{\omega}_k$ requires to redefine some quantities.
In particular, we define the functions $\bar{D}$ and $\widehat{D}$ from $\mathbb{R}^K$ to $\mathbb{R}$, and detail their respective derivatives (here $\bu = (u_1, \ldots, u_K)$ is a generic point in $\mathbb{R}_+^K$).
\begin{align}
\bar{D}(\bu) &= \int \log\left[ \sum_{l=1}^K e^{u_l} \omega_l(z)\right]d\bar{p}(z) - \sum_{l=1}^K \lambda_l u_l\,,\nonumber\\[0.2cm]
\widehat{D}(\bu) &= \int \log\left[ \sum_{l=1}^K e^{u_l} \hat{\omega}_l(z)\right]d\hat{p}(z) - \sum_{l=1}^K \lambda_l u_l\,,\label{eq:min_D}\\[0.2cm]
\left[\bar{D}'(\bu)\right]_k &= \int \frac{e^{u_k} \omega_k(z)}{\sum_{l=1}^K e^{u_l} \omega_l(z)}d\bar{p}(z) - \lambda_k\,,\nonumber\\[0.2cm]
\left[\widehat{D}'(\bu)\right]_k &= \int \frac{e^{u_k} \hat{\omega}_k(z)}{\sum_{l=1}^K e^{u_l} \hat{\omega}_l(z)}d\hat{p}(z) - \lambda_k\,,\nonumber\\[0.2cm]
\left[\bar{D}''(\bu)\right]_{k, k'} &= \int \left[\frac{e^{u_k} \omega_k(z) \delta_{kk'}}{\sum_{l=1}^K e^{u_l} \omega_l(z)} - \frac{e^{u_k} \omega_k(z) e^{u_{k'}} \omega_{k'}(z)}{\left(\sum_{l=1}^K e^{u_l} \omega_l(z)\right)^2}\right]d\bar{p}(z)\,,\label{eq:hessian}\\[0.2cm]
\left[\widehat{D}''(\bu)\right]_{k, k'} &= \int \left[\frac{e^{u_k} \hat{\omega}_k(z) \delta_{kk'}}{\sum_{l=1}^K e^{u_l} \hat{\omega}_l(z)} - \frac{e^{u_k} \hat{\omega}_k(z) e^{u_{k'}} \hat{\omega}_{k'}(z)}{\left(\sum_{l=1}^K e^{u_l} \hat{\omega}_l(z)\right)^2}\right]d\hat{p}(z)\,.\nonumber
\end{align}
The difference with the original proof lies in the control of
\[\sup_{\bu, k,k'} ~ \left|\big[\widehat{D}''(\bu)\big]_{k,k'} - \big[\bar{D}(\bu)''\big]_{k,k'}\right|\,,\]
since we now have to account for the difference between the $\omega_k$ and $\hat{\omega}_k$.
For $\bu \in \mathbb{R}_+^K$ and $z \in \mathcal{Z}$, let $q_k(z) = e^{u_k}\omega_k(z) / \big(\sum_l e^{u_l} \omega_l(z)\big)$, and $\hat{q}_k(z) = e^{u_k}\hat{\omega}_k(z) / \big(\sum_l e^{u_l} \hat{\omega}_l(z)\big)$.
Note that since $u_k \ge 0$ we have for any $z$ that $\sum_l e^{u_l} \omega_l(z) \ge \sum_l \omega_l(z) \ge \varepsilon$, and similarly $\sum_l e^{u_l} \hat{\omega}_l(z) \ge \varepsilon/2$ by \eqref{eq:lower_omega}.
Then we have
\begin{align*}
|\hat{q}_k(z) - q_k(z)| &= \left|\frac{e^{u_k}\hat{\omega}_k(z) \big(\sum_l e^{u_l} \omega_l(z)\big) - e^{u_k}\omega_k(z)\big(\sum_l e^{u_l} \hat{\omega}_l(z)\big)}{\big(\sum_l e^{u_l} \omega_l(z)\big)\big(\sum_l e^{u_l} \hat{\omega}_l(z)\big)}\right|\\
&\le \frac{2}{\varepsilon^2} \left| e^{u_k}\hat{\omega}_k(z) \Big(\sum_l e^{u_l} \omega_l(z)\Big) - e^{u_k}\omega_k(z)\Big(\sum_l e^{u_l} \omega_l(z)\Big) \right|\\
&~~+\frac{2}{\varepsilon^2} \left| e^{u_k}\omega_k(z) \Big(\sum_l e^{u_l} \omega_l(z)\Big) - e^{u_k}\omega_k(z)\Big(\sum_l e^{u_l} \hat{\omega}_l(z)\Big) \right|\\
&\le \frac{2Ke^{2U}}{\varepsilon^2}|\hat{\omega}_k(z) - \omega_k(z)| + \frac{2e^{2U}}{\varepsilon^2} \sum_l |\hat{\omega}_l(z) - \omega_l(z)|\\
&\le \frac{4KC_\omega e^{2U}}{\varepsilon^2\sqrt{n}}\,,
\end{align*}
and
\[
|\hat{q}_k(z) \hat{q}_{k'}(z) - q_k(z)q_{k'}(z)| \le \hat{q}_k(z) |\hat{q}_{k'}(z) - q_{k'}(z)| + q_{k'}(z) |\hat{q}_k(z) - q_k(z)| \le \frac{8KC_\omega e^{2U}}{\varepsilon^2\sqrt{n}}\,,
\]
such that
\begin{align}
\Big|\big[\widehat{D}''(\bu)\big]_{k,k'} &- \big[\bar{D}(\bu)''\big]_{k,k'}\Big|\nonumber\\
&= \left|\int \Big(\hat{q}_k(z)\delta_{kk'} - \hat{q}_k(z)\hat{q}_{k'}(z)\Big) d\hat{p}(z) - \int \Big(q_k(z)\delta_{kk'} - q_k(z)q_{k'}(z)\Big) d\bar{p}(z)\right|\nonumber\\
&\le \int \left|\Big(\hat{q}_k(z)\delta_{kk'} - \hat{q}_k(z)\hat{q}_{k'}(z)\Big) - \Big(q_k(z)\delta_{kk'} - q_k(z)q_{k'}(z)\Big) \right| d\hat{p}(z)\nonumber\\
&~~+ \int \left|\Big(q_k(z)\delta_{kk'} - q_k(z)q_{k'}(z)\Big) \right| (d\hat{p} - d\bar{p})(z)\label{eq:foo}\\
&\le \frac{12KC_\omega e^{2U}}{\varepsilon^2\sqrt{n}} + t\nonumber
\end{align}
with probability at least $1-2K\exp\left(-\frac{\underline{\lambda}nt^2}{2}\right)$ where we have used Corollary 2 from LC19 to bound the second term in \eqref{eq:foo}.
The rest of the proof is similar to that of Proposition 4 in LC19, except that we have $12KC_\omega e^{2U}/\varepsilon^2$ instead of $C_\lambda K$ as constant.
\medskip

\par{\bf Adapting Proposition 5 from LC19.}
Similarly to Proposition 4, the difference with LC19 comes from the fact $\widehat{D}'(\bu)$ and $\bar{D}'(\bu)$ do not differ because of the difference between $\hat{\lambda}_k$ and $\lambda_k$ but because of that between $\hat{\omega}_k$ and $\omega_k$.
Hence, the proposition holds true with $4K^{3/2}C_\omega e^{2U}/(\varepsilon^2\sqrt{n})$ instead of $2C_\lambda K^{3/2}/\sqrt{n}$ as first term.
\medskip

\par{\bf Adapting Proposition 2 from LC19.}
The proposition is proved by using the intermediate results established above.
Again, only the constant factors change.
\medskip

\par{\bf Adapting Proposition 3 from LC19.}
Again, we can follow the path from LC19, using \eqref{eq:lower_omega} when necessary.
\medskip

\par{\bf Adapting Theorem 1 from LC19.}
We follow the same path, using \eqref{eq:lower_omega} when necessary.
Only the constant factors may change due to the adaptation of the proof.
\medskip

\par{\bf Proof of \Cref{thm:approx}.}
Once the adaptation of Theorem 1 in LC19 is proved, one may just use the classical argument stipulating that
\[
\mathcal{R}(\tilde{h}_\mathrm{approx}) - \mathcal{R}(\tilde{h}^*) \le 2 \sup_{h \in \mathcal{H}} \left|\frac{1}{n}\sum_{k=1}^K\sum_{i=1}^{n_k} \pi_{k, i}\ell\big(h(x_i^{(k)}), y_i^{(k)}\big) - \mathcal{R}(h)\right|\,.
\]
The right-hand side is precisely the quantity bounded in [LC19, Theorem 1], such that all constants are just multiplied by $2$.\qed

\section{Computation of the \texorpdfstring{$\hat{W}_k$}{W}}
\label{apx:W}

In this appendix, we provide the technical details about the computation of the $\hat{W}_k$.
Recall that the $\hat{W}_k$ are empirical estimators of the $\Omega_k$, that are meant to be plugged into \eqref{eq:debias} to produce an unbiased estimate of $\ptest$.
For all $k \in \{1, \ldots, K\}$, we have
\[
\Omega_k = \int \omega_k(z)d\ptest(z)\,, \qquad \text{and} \qquad \ptest = \left(\sum_{l=1}^K \frac{\lambda_l\omega_l}{\Omega_l}\right)^{-1}\bar{p}\,.
\]
Plugging the definition of $\ptest$ into $\Omega_k$, we obtain that for all $k \in \{1, \ldots, K\}$
\[
\Omega_k = \int \frac{\omega_k(z)}{\sum_{l=1}^K \frac{\lambda_l\omega_l}{\Omega_l}}d\bar{p}\,.
\]
\smallskip
In other words, $\bm{\Omega} = (\Omega_1, \ldots, \Omega_K) \in \mathbb{R}^K$ is a solution to the system of $K$ equations
\[
\forall k \in \{1, \ldots, K\}, \quad \Gamma_k(\bW) = 1\, \qquad \text{where} \qquad \Gamma_k(\bW) = \frac{1}{W_k}  \int  \frac{\omega_k(z)}{\sum_{l=1}^K \frac{\lambda_l\omega_l(z)}{W_l}}d\bar{p}(z)\,.
\]
Therefore, a natural estimator $\hat{\bW} = (\hat{W}_1, \ldots, \hat{W}_K)$ of $\bOmega$ is given by the solution to the system of equations
\begin{equation}\label{eq:sys_emp}
\forall k \in \{1, \ldots, K\}, \qquad \hat{\Gamma}_k(\bW) = 1\,,
\end{equation}
where $\hat{\Gamma}_k$ is the empirical counterpart of $\Gamma_k$ (i.e., with $\hat{p}$ instead of $\bar{p}$).
Note that $\Gamma_k$ (and similarly $\hat{\Gamma}_k$) is homogeneous of degree $0$, such that an infinite number of solutions might exist.
To ensure uniqueness, one can enforce $\hat{W}_K = 1$ and solve the system composed of the first $K-1$ equations only.
This convention has no impact on $\tilde{p}$, as it is equivalent to plug $\hat{\bW}$ or $t\hat{\bW}$ into \Cref{eq:debias}.
The remaining question is: \textit{how to solve system \eqref{eq:sys_emp}?}
Let $u_k = \log(\lambda_k / W_k)$.
We can rewrite system \eqref{eq:sys_emp} as
\begin{equation}\label{eq:sys_u}
\forall k \in \{1, \ldots, K\}, \qquad \int \frac{e^{u_k} \omega_k(z)}{\sum_{l=1}^K e^{u_l} \omega_l(z)}d\hat{p}(z) - \lambda_k = 0\,.
\end{equation}
Now, the left-hand side of \eqref{eq:sys_u} can be interpreted as the $k^\text{th}$ component of the gradient of the function $\widehat{D}$ defined in \eqref{eq:min_D}.
Thus, solving system \eqref{eq:sys_emp} is equivalent to solving system \eqref{eq:sys_u}, and amounts to maximizing over $\mathbb{R}^K$ the function $\widehat{D}$, which has been shown in \cite{gill1988} to be strongly convex if the distributions overlap sufficiently.
A standard Gradient Descent strategy (whose gradient is given by \Cref{eq:sys_u}) can thus be used, and converges towards the unique minimum.
Recall also that
\[
\hat{p}(z) = \frac{1}{n} \sum_{k=1}^K\sum_{i=1}^{n_k} \mathbb{I}\big\{z = z_i^{(k)}\big\}\,,
\]
so that $\widehat{D}$ rewrites
\[
\widehat{D}(u) = \frac{1}{n} \sum_{k=1}^K\sum_{i=1}^{n_k} \left( \log \left[ \sum_{l=1}^K e^{u_l} \omega_l\big(z_k^{(i)}\big) \right] - \sum_{l=1}^K \lambda_l u_l\right)\,.
\]
The function $\widehat{D}$ is thus separable in the observations, and a stochastic or minibatch variant of Gradient Descent can be used to reduce the computational cost.

\end{document}